# Role of Uncertainty in Model Development and Control Design for a Manufacturing Process

**Rongfei Li [1] and Francis Assadian [2]**

[1] University of California, Davis; rfli@ucdavis.edu
[2] University of California, Davis; fassadian@ucdavis.edu

## Abstract

The use of robotic technology has drastically increased in manufacturing in the 21st century. But by utilizing their sensory cues, humans still outperform machines, especially in the micro scale manufacturing, which requires high-precision robot manipulators. These sensory cues naturally compensate for high level of uncertainties that exist in the manufacturing environment. Uncertainties in performing manufacturing tasks may come from measurement noise, model inaccuracy, joint compliance (e.g., elasticity) etc. Although advanced metrology sensors and high-precision microprocessors, which are utilized in nowadays robots, have compensated for many structural and dynamic errors in robot positioning, but a well-designed control algorithm still works as a comparable and cheaper alternative to reduce uncertainties in automated manufacturing. Our work illustrates that a multi-robot control system can reduce various uncertainties to a great amount.

## 1. Introduction

It is believed that the rapid emergence of Robotic technology in industry, and specifically in manufacturing, in the 21st century, will have positive impacts in many aspects of our lives. We have already seen many applications of this technology in macro scale, such as pick and place task [1]. However, there are still applications where humans outperform machines, especially in the micro scale manufacturing, which requires high-precision robot manipulators.

Accurate positioning of robot arms is very important in automated manufacturing field. Over past several decades, we have seen great strides in the technology for accurate positioning robots. We have seen researchers have tried to implement add-on features such as real-time microprocessors, high precision motors, zero backlash gear set, advanced metrology sensors and so on in today's robots. Indeed, they have compensated many structural and dynamic errors in robot positioning [2]. However, those add-on features are usually very expensive and unnecessarily increase the cost during the manufacturing process. Robotic systems that employ a well- designed sensor-based control strategies can reduce the cost and simultaneously obtain robustness against disturbances and imprecisions from sensors or modeling.

The process of fastening and unfastening a screw is a mundane but a challenging task in the automated manufacturing. We found recent research on this topic only focuses on how robots should generate push/pull force on a driver [3]. The axial forces and torques are first measured through sensors and then controlled to imitate human approach of fastening and unfastening by applying similar amount of axial forces and torques. This approach only considers tactile sensing, however, human beings, also use the information from visual sensing to help with this task. To replicate visual sensing in robots, for example, a camera system could be utilized to make sure a tool is at the right pose (correct orientation and location where head of bolt and tail of driver coincide). A visual system that can provide an accurate and repeatable positional tracking of the tool becomes significantly important and useful not only in this type of an application but also in many other applications of the automated manufacturing [4].

In this work, we have designed a multi-robotic control system that simulates the positioning process for fastening and unfastening applications and have examined its robustness against various uncertainties, which may occur, in this process. This control system is a visual servoing system where a camera is mounted on a robot arm manipulator and provides vision data for the motion control of a second robot manipulator with a tool. Both the Position-Based Visual Servoing (PBVS) and the Image-Based Visual Servoing (IBVS) systems have been thoroughly investigated in [5-8]. However, in these related work, in the visual servoing domain, the development of the outer-loop controller is usually achieved with the PID controller, or its simplified variations based only on a kinematic model of camera [5,8]. One improvement in this work is to use Youla robust control design technique [9] that includes both kinematics and dynamics in the model development stage. The increase in the model fidelity for the control design can positively influence the precision of the feature estimation and the control system stability for the high-speed tasks. Benefits of our design are discussed in more details in the following sections.

Position control algorithms for both the visual and the tool manipulation systems are discussed in this Chapter. Especially, a combination of a feedforward and a feedback control architecture has been designed for the tool manipulation system, which enables the tool to move fast to a desired location with a high precision in its final pose. Simulation results for the Single Input Single Output (SISO) case in various scenarios are presented and furthermore, the robustness to various noise sources in this manufacturing process are examined.

## 2. Literature review

We have seen many efforts been made to improve the positioning accuracy of robotic systems in the past few decades. An effective way to reduce the amount of inaccuracy is to measure it with sensors and compensate is through feedback control loop. Many metrology techniques have been investigated and applied for different kinds of data capturing. Among them, three methods have gained the most popularity in the recent research, namely, vision-based methods, tactile-based methods, and vision-tactile integrated methods. In this section, we will briefly review those approaches and their applications.

## 2.1 Vision-based methods

The vision-based methods have been widely developed in the recent years and used to determine position and orientation of target objects in robotic systems. Zhu et al. discussed Abbe errors in a 2D vision system for robotic drilling. Four laser displacement sensors were used to improve the accuracy of the vison-based measurement system [5]. Liu et al. proposed a visual servoing method for positioning in aircraft digital assembly [6,7]. With the measurements from two CCD cameras and four distances sensors, the proposed method can accurately align the positioner's ball-socket with the ball-head fixed on the aircraft structures in a finite time.

In addition to mentioned applications, we have seen many contributions to the vision-based methods in robotic manipulation. However, most of those researchers focused on success rate of grasping on end-effector without enough analysis on the positioning accuracy. Du et al. published a study for the robotic grasp detection by visually localizing the object and estimating its pose [8]. Avigal et al. proposed a 6-DoFs grasp planning method using fast 3D reconstruction and grasp quality convolutional neural network (CNN) [9]. Wu et al. proposed an end-to-end solution for visual learning [10].

## 2.2 Tactile-based methods

In addition, with the development of tactile sensors in the last few years, we have seen more and more focus on tactile-based methods in robotic positioning domain. The tactile sensors can show contact states of the end-effector and the object in robotic manipulations. The contact state can be used to determines objects' relative orientations and positions with respective to the gripper. Li et al. designed a tactile sensor of GelSight and generated tactile maps for different poses of a small object in the gripper [10]. He studied the localization and control manipulation for a specific USB connector insertion task. Dong et al. studied the tactile-based insertion task for dense box packing with two GelSlim fingers which are used to estimate object's pose error based on neural network [11]. Furthermore, Hogan et al. developed a tacile-based feedback loop in order to control a dual-palm robotic system for dexterous manipulations [12]. Those tactile-based methods can only realize relative accurate positioning of the tool with the end-effector, but the positioning of the robot manipulator itself is not addressed.

## 2.3 Vision-tactile integrated methods

Vision sensing can provide more environment information with a wide measurement range, while tactile sensing can provide more detailed information in robotic manipulations. Therefore, the vision–tactile integrated methods came into being. Fazeli et al. proposed a hierarchical learning method for complex manipulation skills with multisensory fusion in seeing and touching [12]. Gregorio et al. developed a manipulation system for automatic electric wires insertion performed by an industrial robot with a camera and tactile sensors implemented on a commercial gripper [13].

According to the analysis, all the integrated sensory applications have achieved accurate robotic manipulation tasks such as insertion and their performances have been verified in experiments. However, the error space in those references is usually small and none of them has considered all the translational and rotational errors in 6 DoFs. Moreover, tactile-based or vision-tactile integrated methods will increase expense of massive manufacturing because tactile sensors are more expensive to purchase and maintain compared to visual sensors. Based on that, our work is to explore the capability of the vison-based methods and design a new method to improve the accuracy of positioning in the robot manipulation system.

## 2. Uncertainty sources

Uncertainties in automated manufacturing can originate from different sources. We can divide these uncertainties into two categories: sensor measurement noise, and dynamic and kinematic modeling errors from both the measurement system and the robot manipulators. This section briefly reviews each uncertainty source including the proposed methods for reducing these uncertainties.

### 2.1 A brief overview of a stereo camera model and its calibration

A camera model (i.e., the pin-hole model [10]) has been adopted in the visual servoing techniques to generate an interaction matrix [5]. The object depth, the distance between a point on the object and the camera center as illustrated in Fig 1, needs to be either estimated or approximated by an interaction matrix [5]. One of the methods is to directly measure the depth by a stereo (binocular) camera with the use of two image planes [11].

As shown in Figure 1, two identical cameras are separated by a baseline distance b. An object point, $\boldsymbol{P}^C = [X^C, Y^C, Z^C]^T$, which is measured in the camera frame, at the baseline center, is projected to two parallel virtual image planes, and each plane is located between each optical center ($\boldsymbol{C}_l$ or $\boldsymbol{C}_R$) and the object point $\boldsymbol{P}^C$. The intrinsic camera parameters relate the coordinates of the object point in the camera frame and its corresponding image coordinates $\boldsymbol{p} = (u, v)$ on each of the image plane with

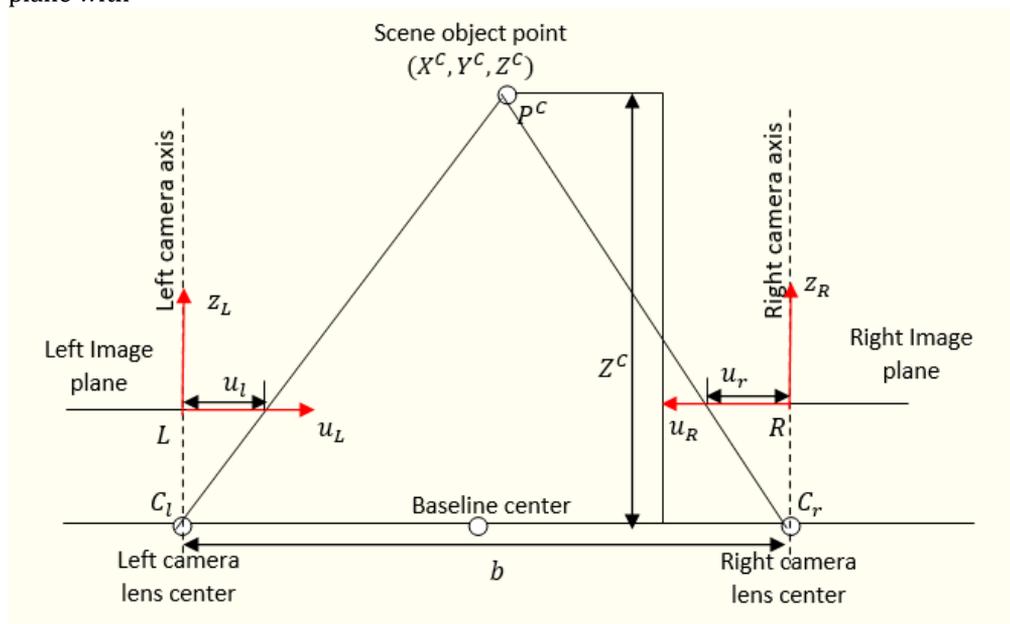

Figure 1. The projection of a scene object on the stereo camera's image planes.

an exact mathematical relationship. This relationship is given by:

**Note:** $v$ coordinate on each image plane is not shown in the plot but is measured along the axis that is perpendicular to and point out of the plot.

$$\begin{bmatrix} u_l \\ v_l \\ 1 \end{bmatrix} = \frac{1}{Z^C} \begin{bmatrix} f_u & s_c & u_0 \\ 0 & f_v & v_0 \\ 0 & 0 & 1 \end{bmatrix} \begin{bmatrix} X^C \\ Y^C \\ Z^C \end{bmatrix} - \frac{b}{2Z^C} \begin{bmatrix} f_u \\ 0 \\ 0 \end{bmatrix} \qquad (1)$$

$$\begin{bmatrix} u_r \\ v_r \\ 1 \end{bmatrix} = \frac{1}{Z^C} \begin{bmatrix} f_u & s_c & u_0 \\ 0 & f_v & v_0 \\ 0 & 0 & 1 \end{bmatrix} \begin{bmatrix} X^C \\ Y^C \\ Z^C \end{bmatrix} + \frac{b}{2Z^C} \begin{bmatrix} f_u \\ 0 \\ 0 \end{bmatrix} \qquad (2)$$

Where, $f_u$ and $f_v$ are the horizontal and the vertical focal lengths, and, $s_c$ is a skew coefficient. In most cases, $f_u$ and $f_v$ are different if the image horizontal and vertical axes are not perpendicular. In order not to have negative pixel coordinates, the origin of the image plane will be usually chosen at the upper left corner instead of the center. $u_0$ and $v_0$ describe the coordinate offsets. The camera model uncertainties may arise from the estimation of those camera intrinsic parameter values. The camera calibration can be used to precisely estimate these values.

The stereo camera calibration has been well studied in [12-15]. As summarized in [12], the calibration method can be divided into two broad categories: the photogrammetric calibration and the self-calibration. In the photogrammetric calibration [13], the camera is calibrated by observing a calibration object whose geometry is well known in the 3D space. These methods are very accurate but require an expensive apparatus and elaborate setups [12]. The self-calibration [12,14,15] is performed by finding the equivalences between the captured images of a static scene from different perspectives. Although cheap and flexible, these methods are not always reliable [12]. The author in [12] proposed a new self-calibration technique that observe planar pattern at different orientations and showed improved results.

## 2.2 A brief overview of the robot manipulator model and its calibration

In this work, we consider the elbow manipulators [18] with the spherical wrist in the multi-robot system to move an end-effector freely in 6 degrees of freedoms (dofs). This model of the robot has six links with three for the arms and the other three for the wrist. The robot arms freely move the end effector to any position in the reachable space with 3 dofs while the robot spherical wrists allow the end effector to orient in any directions with another 3 dofs. For the elbow manipulators, a joint is connected between each two adjacent links and there are in total six convolutional joints. The specific industry model of this type is ABB IRB 4600 [17].

A commonly used convention for selecting and generating the reference frames in the robotic applications is the Denavit-Hartenberg convention (or D-H convention) [18]. Suppose each link is attached to a Cartesian coordinate frame, $O_i X_i Y_i Z_i$. In this convention, each homogeneous transformation matrix $A_i$ (from frame $i-1$ to frame $i$) can be represented as a product of four basic transformations:

$$A_i = Rot_{z,q_i} Trans_{z,d_i} Trans_{x,a_i} Rot_{x,\alpha_i} \qquad (3)$$

$$= \begin{bmatrix} c_{q_i} & -s_{q_i}c_{\alpha_i} & s_{q_i}s_{\alpha_i} & a_ic_{q_i} \\ s_{q_i} & c_{q_i}c_{\alpha_i} & -c_{q_i}s_{\alpha_i} & a_is_{q_i} \\ 0 & s_{\alpha_i} & c_{\alpha_i} & d_i \\ 0 & 0 & 0 & 1 \end{bmatrix}$$

**Note:** $c_{q_i} \equiv \cos(q_i)$, $c_{\alpha_i} \equiv \cos(\alpha_i)$, $s_{q_i} \equiv \sin(q_i)$, $s_{\alpha_i} \equiv \sin(\alpha_i)$.

Where $q_i$, $a_i$, $\alpha_i$ and $d_i$ are parameters of link $i$ and joint $i$, $a_i$ is the link length, $q_i$ is the rotational angle, $\alpha_i$ is the twist angle and $d_i$ is the offset length between the previous $(i-1)^{th}$ and the current $i^{th}$ robot links. The quantities of each parameter in (3) are calculated based on the steps in [16].

We can generate the transformation matrix from the base frame $O_0X_0Y_0Z_0$ ($P^0$) to the end-effector frame $O_6X_6Y_6Z_6$ ($P^6$):

$$T_6^0 = A_1^0 A_2^1 A_3^2 A_4^3 A_5^4 A_6^5 \tag{4}$$

If any point with respect to the end effector frame $P^6$ is known, we can calculate its coordinate with respect to the base frame $P^0$ as:

$$P^0 = T_6^0 P^6 \tag{5}$$

In addition, the transformation from the base frame $P^0$ to the end-effector frame $P^6$ can be derived:

$$T_0^6 = (T_6^0)^{-1} \tag{6}$$

which is used to generate the image coordinates of a point captured by a camera with its center attached to the end effector, from the 3D coordinates of a point in the base frame.

Eq. (4) shows that the position of the end-effector $P^{end}$ (where $P^{end}$ is the origin of the end-effector frame $P^6$) is a function of all the joint angles $q = [q_i | i \in 1,2,3,4,5,6]$ and the parameters $Pa = [a_i, \alpha_i, d_i | i \in 1,2,3,4,5,6]$:

$$P^{end} = \mathcal{F}(q, Pa) \tag{7}$$

Eq. (7) describes the forward kinematic model of the robot manipulator, which could be utilized to calculate the position of the end effector from the joint angles and the robot parameters. The inverse process is called the inverse kinematic, which the joint angles can be computed from the position and the parameters. The estimation of the robot parameters $Pa$ determines the accuracy of the kinematic models of the robot manipulators.

The paper [19] provides a good summary of the current robot calibration methods. The author of [20] states that over 90% of the position errors are due to the errors in the robot zero position (the kinematic parameter errors). As a result, most researchers focus on the kinematic robot calibration (or level 2 calibration [19]) to enhance the robot absolute positioning accuracy [21-24]. Generally, the kinematic model-based calibration involves four sequential steps: Modeling, Measurement, Identification, Correction. Modeling is a development of a mathematical model of the geometry and the robot motion. The most popular one is D-H convention [18] and other alternatives include S-model [25] and zero-reference model [26]. At the measurement step, the absolute position of the end-effector is measured from the sensors, e.g., the acoustic sensors [25], the visual sensors [22], etc. In the identification step, the parameter errors for the robot are identified by minimizing the residual position errors with different techniques [27, 28]. This final step is to implement the new model with the corrected parameters.

On the other hand, the non-kinematic calibration modeling (level 3 calibration [19]) [27, 29], which includes the dynamic factors such as the joint and the link flexibility in the calibration, increase accuracy of the robot calibration, but complicates the mathematical functions that govern the parameters relationship.

## 2.3 The Image Averaging Techniques for Denoising

The image noises are inevitably introduced in the image processing. Several image denoising techniques have been proposed so far. A good noise removal algorithm ought to remove as much noise as possible while safeguarding the edges. The Gaussian white noise has been dealt with using the spatial filters, e.g., the Gaussian filter, the Mean filter and the Wiener filter [30]. The noise reduction using the wavelet methods [31,32] have benefits of keeping more useful details but at the expense of the computational complexity. However, depending on the selected wavelet methods, the filters that operate in the wavelet domain still filter out (or blur) some important high frequency useful information of the original image, even though more edges are preserved with the wavelet method when comparing with the spatial filter approaches.

All the aforementioned methods present ways to reduce noise in the image processing starting from a noisy image. We can approach this problem with the multiple noisy images taken from the same perspective. Assuming the same perspective ensures the same environmental conditions (illumination, temperature, etc.) that affect the image noise level. Given the same conditions, an image noise level taken at a particular time should be very similar to another image taken at a different time. This redundancy can be used for the purpose of improving image precision estimation in the presence of noise. The method that uses this redundancy to reduce noise is called signal averaging (or the image averaging in the application of the image processing) [33]. The image averaging has a natural advantage of retaining all the image details as well as reducing the unwanted noises, given that all the images for the averaging technique are taken from the same perspective. The robot's rigid end effector that holds a camera minimizes shaking and drift when shooting pictures. Furthermore, in the denoising process, the precise estimations require that the original image details to be retained. Considering these previous statements, we decided to choose image averaging over all other denoising techniques in this work.

The image averaging technique is illustrated in Figure 2. Assume a random, unbiased noise signal, and in addition, assume that this noise signal is completely uncorrelated with the image signal itself. As noisy images are averaged, the original true image is kept the same and the magnitude of the noise signal is compressed thus improving the signal-to-noise ratio. In Figure 2, we generated two random signals with the same standard deviation, and they are respectively represented by the blue and the red lines. The black line is the average of the two signals, whose magnitude is significantly decreased compared to each of the original signal. In general, we can come up with a mathematical relationship between the noise level reduction and the sample size for averaging. Assume we have $N$ numbers of Gaussian white noise samples with the standard deviation $\sigma$. Each sample is denoted as $z_i$, where $i$ represents $i^{th}$ sample signal. Therefore, we can acquire that:

$$var(z_i) = E(z_i^2) = \sigma \qquad (8)$$

where $E(\cdot)$ is the expectation value and $\sigma$ is the standard deviation of the noise signal. By averaging the $N$ Gaussian white noise signals, we can write:

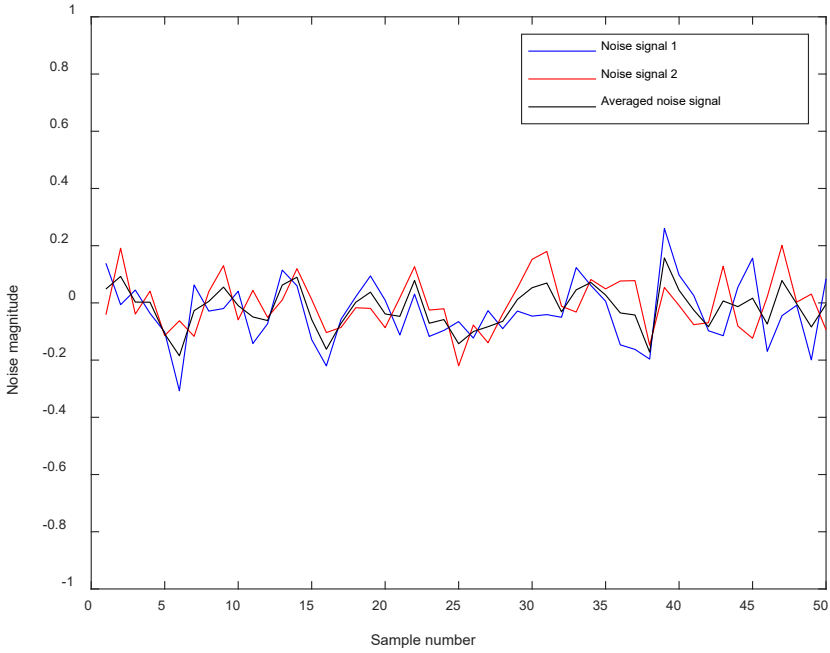

Figure 2. An example of a noise level reduction by Image averaging.

$$var(z_{\text{avg}}) = var(\frac{1}{N}\sum_{i=1}^{N} z_i) = \frac{1}{N^2} N\sigma^2 = \frac{1}{N}\sigma^2 = (\frac{1}{\sqrt{N}}\sigma)^2 \qquad (9)$$

where $\mathbf{z_{avg}}$ is the average of the N noise signals. Eq. (9) demonstrates that a total number of N samples are required to reduce the signal noise level by $\sqrt{N}$. Since our goal is to reduce the image noise to within a fixed threshold (a constant number expressed as a standard deviation), a smaller variation in the original image requires much less samples to make an equivalent noise-reduction estimation. Thus, it is worthwhile for the camera to move around, rather than being stationary, in order to find the best locations where the image noise level estimation is small. In general, we can reduce the noise level as much as needed by taking more samples.

## 2.4 The dynamic errors and its modeling in the feedback control loop

The dynamic errors in a robot manipulator consist of any joint non-static errors. Among these error sources, which have the most significant effect on the robot position control accuracy, are the deviations between the actual joint rotation with its measured value from the unmodeled dynamic uncertainties, such as backlash, friction, compliance due to gears' elimination, joint or link flexibility and thermal effects. As discussed in Section 2.2, we can account for all these errors in the dynamic modeling step by developing a high-fidelity dynamic model, where all these parameter values could be identified through calibration. An easier way is to regard all these dynamic errors as disturbances to a manipulator control system. The control system is able to make the plant output track the desired input (the reference signal) and while simultaneously, it rejects these disturbances. The design of the robot manipulators control systems and the demonstrations of the capability of these feedback loops to reject these aforementioned dynamic errors are discussed in more detail in the later sections.

## 3. A brief overview of the classical IBVS architecture

The control of the visual system, the hand-in-eye camera configuration [34], is discussed in this section. The general visual servoing problem can be divided into two categories: PBVS and IBVS [5]. This work focuses on the IBVS structure. Figure 3 shows the control block diagram for a classical IBVS architecture. In Figure 3, $s = [u, v]^T$ is the image feature position vector, $s^* = [u^*, v^*]^T$ is the target image feature position vector, and their difference $e = s^* - s$ is the error vector. $L_e$ is the so-called interaction matrix [5], which is a 2-by-6 matrix, and relates the time derivative of the image feature $s$ to the spatial velocity of the camera $V_c$, a column vector of six-elements, by the following:

$$\dot{s} = L_e V_c \tag{10}$$

We can design a proportional controller to force the error to exponentially converge to zero, i.e.:

$$\dot{e} = -ke, k > 0 \tag{11}$$

Suppose the target image feature is a constant; that is $\dot{s}^* = 0$, hence, we can derive from Eq. (11):

$$\dot{e} = \dot{s}^* - \dot{s} = -L_e V_c \tag{12}$$

From Eqs. (11) and (12), one will be able to obtain:

$$V_c = k L_e^+ e \tag{13}$$

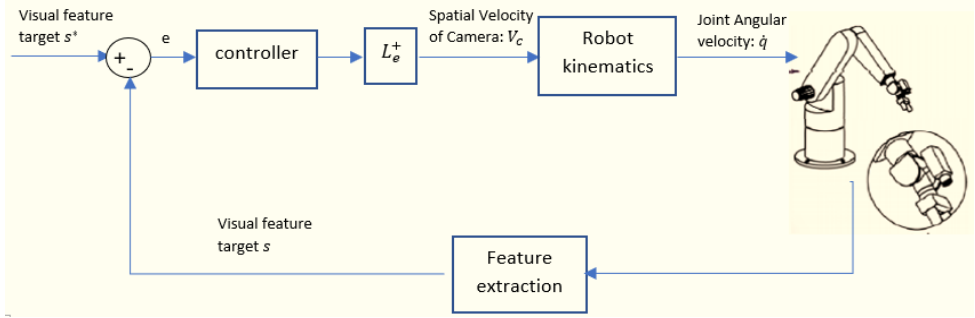

Figure 3. The block diagram of the classical IBVS control architecture.

where $L_e^+$ is the pseudoinverse of $L_e$. The derivation of the interaction matrix with a monocular camera is further explained next. We assume a point with the three-dimension (3D) coordinates in the camera frame is given as $P^C = [X^C, Y^C, Z^C]$. We further assume a zero-skew coefficient, i.e., $s_C = 0$ in Eq. 2 with a baseline distance $b = 0$, for a monocular camera model, then the image feature coordinate $s = [u, v]^T$ can be expressed as:

$$\begin{bmatrix} u \\ v \end{bmatrix} = \begin{bmatrix} \dfrac{f_u X^C}{Z^C} + u_0 \\ \dfrac{f_v Y^C}{Z^C} + v_0 \end{bmatrix} \tag{14}$$

Taking the time derivative of Eq. (14), we can obtain:

$$\begin{bmatrix} \dot{u} \\ \dot{v} \end{bmatrix} = \begin{bmatrix} \dfrac{f_u(\dot{X}^C Z^C - \dot{Z}^C X^C)}{(Z^C)^2} \\ \dfrac{f_v(\dot{Y}^C Z^C - \dot{Z}^C Y^C)}{(Z^C)^2} \end{bmatrix} \tag{15}$$

The rigid body motion of a 3D point in the camera model can be derived as:

$$\dot{P}^C = v^C + \omega^C \times P^C \Leftrightarrow \begin{cases} \dot{X}^C = v_X^C - \omega_Y^C Z^C + \omega_z^C Y^C \\ \dot{Y}^C = v_Y^C - \omega_Z^C X^C + \omega_Z^C Z^C \\ \dot{Z}^C = v_Z^C - \omega_X^C Y^C + \omega_y^C X^C \end{cases} \tag{16}$$

Substituting (16) to (15), and rearranging the terms, we obtain:

$$\begin{bmatrix} \dot{u} \\ \dot{v} \end{bmatrix} = \begin{bmatrix} \dfrac{f_u}{Z^C} & 0 & -\dfrac{u-u_0}{Z^C} & -\dfrac{(u-u_0)(v-v_0)}{f_v} & \dfrac{f_u{}^2 + (u-u_0)^2}{f_u} & -\dfrac{f_u(v-v_0)}{f_v} \\ 0 & \dfrac{f_v}{Z^C} & -\dfrac{v-v_0}{Z^C} & -\dfrac{f_v{}^2 + (v-v_0)^2}{f_v} & \dfrac{(u-u_0)(v-v_0)}{f_u} & \dfrac{f_v(u-u_0)}{f_u} \end{bmatrix} \begin{bmatrix} v_X^C \\ v_Y^C \\ v_Z^C \\ \omega_X^C \\ \omega_Y^C \\ \omega_Z^C \end{bmatrix} \tag{17}$$

Eq. (17) can be simply written as:

$$\dot{s} = L_e V_c = L_e \begin{bmatrix} v^C \\ \omega^C \end{bmatrix} \tag{18}$$

Some drawbacks of the classical IBVS are summarized next. To compute the interaction matrix $L_e$ from Eq. (17), the depth $Z^C$ needs to be estimated. This can be usually approximated as either the depth of the initial position or the depth of the target position or their average value [5]. A careless estimation of the depth may lead to a system instability. In addition, the design of the proportional controller is based on, Eq. (10), the camera kinematic relationships, such that there is no dynamics considered in this model. The kinematic model is sufficient for very slow responding system, however, for faster responses, one has to take into account the manipulator dynamics along with the camera model.

In this work, we propose a new controller algorithm, similar to the classical IBVS structure, where the controller is designed with the complete dynamic and kinematic models of the robot manipulator and the camera. Furthermore, this algorithm doesn't require any depth estimation, therefore, it will not be necessary to use the interaction matrix. The development of this new algorithm is presented in sections 6 and 7 of this Chapter.

## 4. The topology of the multi-robotic system for accurate positioning control

In this section, we discuss the proposed control architectures for a multi-robot system, which enables the high-accuracy movement of a tool in various manufacturing scenarios by reducing the process uncertainties. Assuming at the start, all camera and robot manipulators are well calibrated by using one or multiple methods discussed in Sections 2.1 and 2.2, so that the initial camera and robot manipulators parameters are identified. Therefore, in this case, the main uncertainties include the sensor noise and the dynamic modeling errors. Figure 4 shows the overall topology of this multi-robot system.

The multi-robot system is composed of a visual system and a tool manipulation system (Figure 4). In the visual system, a camera is mounted on an elbow robot arm while a tool is held by the end-effector of the robot manipulator arm. The goal of the visual system is to provide precise estimation of the tool pose so that the tool manipulator can control the pose with the guidance from the visual system. Two fiducial markers (green circles or the interest points) are placed on the tool to help the computer to detect the position and the orientation of the tool. The absolute coordinates of the reference points (red circles) are known in the inertial reference frame. The reference points are placed close to the tool's target location (the target interest points) so that the reference points and the target interest points can be captured in the camera frame when the tool gets close to its target pose.

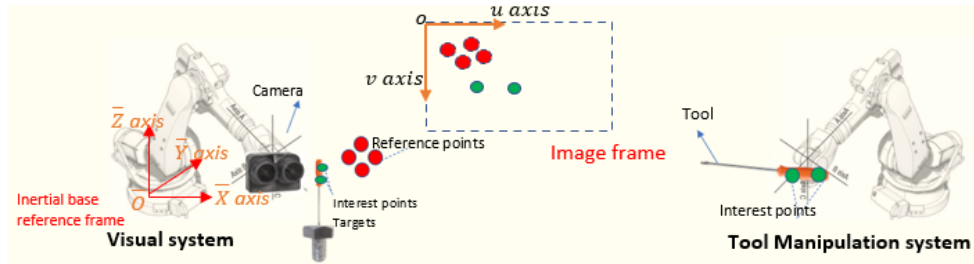

Figure 4. The topology of the multi-robotic system for accurate positioning control.

Four reference points are selected close to each other in the space. In a visual servoing problem, a location in space from which an image was taken can only be uniquely determined by at least four points, 3 points to determine a specific location and one point to determine the orientation. This is a location determination problem (LDP) using image recognition [35]. Therefore, we consider using four reference points to determine the camera pose in the 3D space. However, whenever the camera pose is fixed and known in space, the stereo camera, which can detect the depth, provides the distinct 3D location of a point from the image coordinates.

## 4.1 The multi-robotic system sequential control procedure

The movement control of the robot manipulators is asynchronous in the visual and the tool manipulator systems. A flowchart demonstrating this sequential control process is shown in Figure 5.

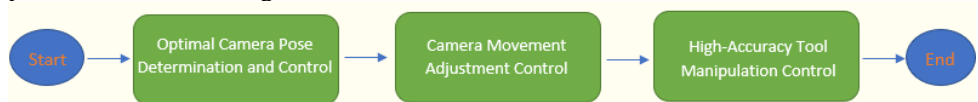

Figure 5. The flow chart of the sequential control procedure.

The first stage consists of the optimal camera pose determination and control. In this stage, the camera moves and searches for an optimal position based on the minimization of a proposed objective function, in this case, the time duration and the energy consumption, while reducing the image noise of the reference points to within an acceptable threshold. In the second stage, the camera movement adjustment control, any uncertainties occur in the movement of camera from the last stage is eliminated by a hand-in-eye visual servoing controller. After the movement adjustment, the camera is kept static and provides precise estimations of the tool position. In the last stage, the high-accuracy tool manipulation control, the tool movement is controlled and guided to the target pose location by a hand-to-eye visual servoing controller. Each control method and architecture are discussed in the sections below.

## 4.2 The optimal camera pose determination process and its control architecture

Figure 6 shows how the optimal pose of the camera is determined from a single picture taken at different perspectives. The uncertainty in the image processing is spatially related. As the camera moves in space, the combined factors (the light conditions, the temperature, etc.) that affect the image processing changes and with these changes, the uncertainty level in the estimation changes accordingly. In this work, we propose to apply image averaging [33] to reduce the uncertainty level in the pose estimation. As discussed in Section 2.3, the number of images required for the averaging increases by a factor of 2 for reducing the uncertainty level by the square root of the same factor. In order to reduce the energy consumption and the time duration in this photo taking process, it is necessary to first determine the location where the image averaging should take place before the camera actually starts to take multiple photos.

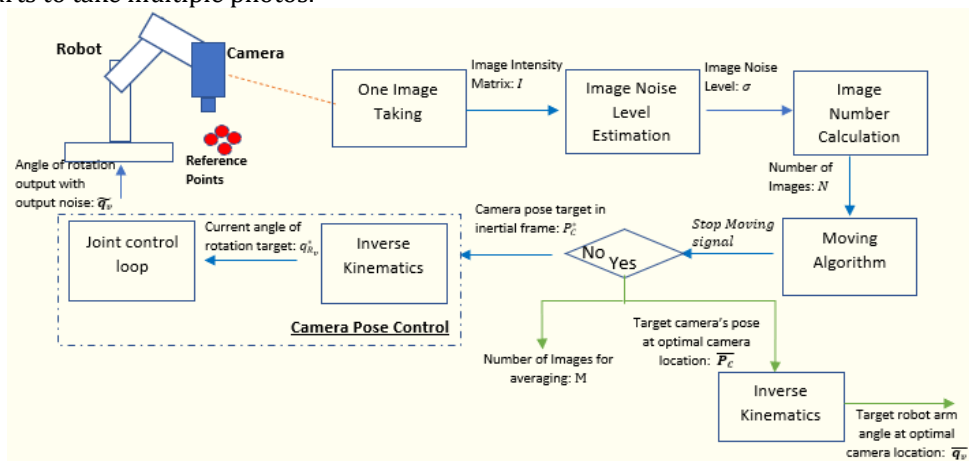

Figure 6. The optimal camera pose determination process and its control architecture.

In Figure 6, in the first stage, the camera takes a single picture. In the second stage, we compute the image intensity matrix $I$ from that photo and then, we estimate the noise level σ across the image by a previously developed algorithm, see [36]. In the third stage, we calculate the uncertainty level from the image noise level and generates the number of images $N$ required to reduce this uncertainty within a prescribed threshold. In the fourth stage, utilizing a moving algorithm, which is designed as a part of this work, the current camera target pose, $P_C^*$ is commanded. In the fifth stage, the camera pose controller guides the camera to the target pose location using the encoders that measure joints rotational angles, $\widetilde{q_v}$. These five stages are repeated until the movement algorithm instructs the camera to stay in the current pose. Then, this current target pose is the optimal pose $\overline{P_C}$ of the camera where the total energy consumption and the time duration is minimized. The output $\overline{q_v}$ is the target joint angles of visual manipulator system at the optimal pose of the camera. If for any reason, such as uncompensated uncertainties, the current pose is not the same as the optimal pose, $\overline{P_C}$, then the camera movement adjustment control, presented in the next section, will reduce this error. In addition, $M$ is the number of pictures needed for the averaging at the optimal pose location of the camera.

## 4.3 The camera movement adjustment control block diagram

We propose a control method with its associated block diagram for the camera movement adjustment as shown in Figure 7. The role of this feedback control is to deal with the errors occurred in the dynamics and the measurements of the previous stage.

In Figure 4, four reference points whose absolute positions are known in the space, are selected close to the tool target pose. The fiducial markers are placed on the reference points, so that their location can be recognized and estimated in a 2D image coordinate frame using computer vision. From the kinematic model of the robot arm and the camera, the image coordinates of the reference points can be calculated online, and those coordinates are used as the targets for a cascaded control loop and are noted as $\overline{p_R}$ in Figure 7. After applying the image averaging technique (Section 2.3), we can obtain a precise estimation of the current position of the reference points and are noted as $\widehat{p_R}$ in Figure 7 in image coordinates from the computer vision. Therefore, any deviation between $\overline{p_R}$ and $\widehat{p_R}$ could be the result of some uncertainties, such as the joint compliances, which are not compensated by the joint control loop as shown in Figure 6. The cascaded controller is similar to the image-based visual servoing scheme (IBVS), as discussed in Section 2.2. The inner-loop control strategy in this part is also very similar to the joint control in the camera pose control in Figure 6, and its control design and simulation results are discussed in Section 5.

Figure 7. The camera movement adjustment control block diagram.

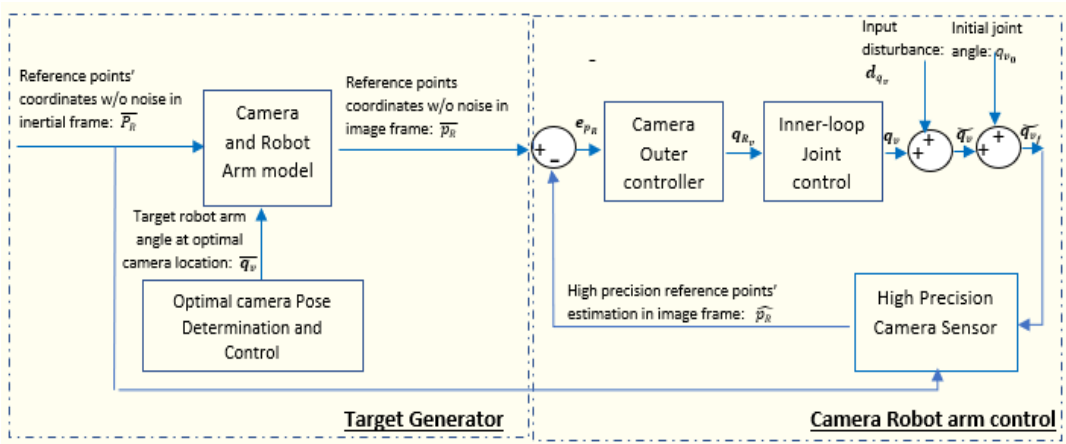

## 4.4 The high-accuracy tool manipulation control block diagram

We propose a control strategy with its associated control block diagram for the tool manipulation system as shown in Figure 8. The control algorithm in this block diagram is a combination of a feedforward and a feedback control.

The feedforward control loop is an open loop which brings the tool as close to the target position as possible in the presence of the input disturbance, $d_{q_m}$. In the inner joint control loop, the noise sources may originate from the low fidelity cheap encoder joint sensors and the dynamic errors from the joint, e.g., compliances. All sources of noise from the joint control loop are combined and modeled as an input disturbance, $d_{q_m}$, to the outer control loop. The outputs of the feedforward are the reference joint angles of rotations, $q_{R_{m_{feedforward}}}$, which are added to the outer feedback controller outputs, $q_{R_{m_{feedback}}}$, and set as the targets for the joint control inner-loop. The function of the forward kinematics is to transform a set of current joint angles of the tool manipulator to the current pose of the tool on the end-effector using a kinematic model of the robot arm.

Movement of the tool can be adjusted with high accuracy by the feedback control loop. The feedback control loop rejects the input disturbance, $d_{q_m}$, and minimizes the error between the tool pose target, $\overline{p_T}$, in the image frame and the high precision estimation tool pose from the camera sensor, $\widehat{p_T}$. The pose in a robot system modeled in Cartesian inertial base frame consists of six degrees of freedom, i.e., three translations and three rotations. Therefore, in order to have a full control of the tool pose, the camera in the feedback control loop requires to measure the image coordinates of at least two interests points on the tool.

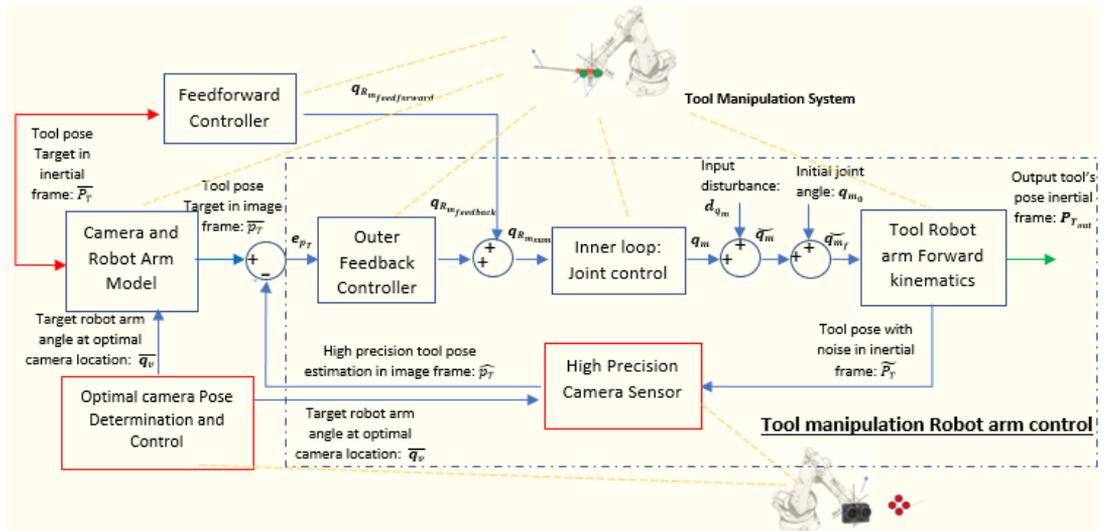

Figure 8. The high-accuracy tool manipulation control block diagram.

The feedforward and feedback controllers work simultaneously to move the tool to the target pose location in the tool manipulation system. The combined target $q_{R_{m_{sum}}}$ are the inputs to the joint control loop so that both controllers manipulate the tool pose. The benefit of designing both feedback and feedforward controls for the manipulation system is to reduce the time duration. If only feedback control is utilized, the pose estimation generated from the visual system requires taking multiple pictures and makes the tool movement very slow. We can divide the task of the tool movement control into two stages. In the first stage, under the action of the feedforward control, the tool moves to an approximate location that is close to the desired destination. In the second stage, the feedback controller moves the tool to the precise target location using the tool pose estimation from the camera. In addition, the camera has a range of view and can only detect the tool and measure its 2D feature as $\widehat{p_T}$ when it is not far away from the target. When the tool is moving from a location that is not in the camera range of view, we must estimate the feature as $\widetilde{p_T}$ until the tool moves into the range of view (this point will be discussed in detail in Section 7). It should be noted that only the feedback controller has the ability to compensate for uncertainties.

This control topology is an analogy to the macro-micro manipulation in the current industry trends where the large-scale robots are used for the approximate positioning, while the small-scale robots are utilized for the precise positioning [2].

### 4.5 The high precision camera sensor model

As shown in Figures 7 and 8, the high precision camera sensor model provides high precision estimations in the feedback loop of the tool manipulation system control and the camera movement adjustment control. The camera robot arm model, which is shown in both Figures 7 and 8, is the target generator that transforms the target in the inertial frame to the target locations in the image frame. The mathematical model of the camera, which is utilized in the visual robot arm and in the feedback loop to generate the required position estimation, has an equivalent Hardware-In-the-Loop (HIL) model as shown in Figure 9.

Figure 9 shows the details of high precision camera model and its equivalent HIL model. The upper configuration is the mathematical model that is used in the simulation to generate image coordinates and to design the outer-loop controller in the robot arm control loop. However, in real application, the lower HIL configuration replaces this mathematical model. In the HIL model, the image processing will make an estimation of the tool pose with high precision.

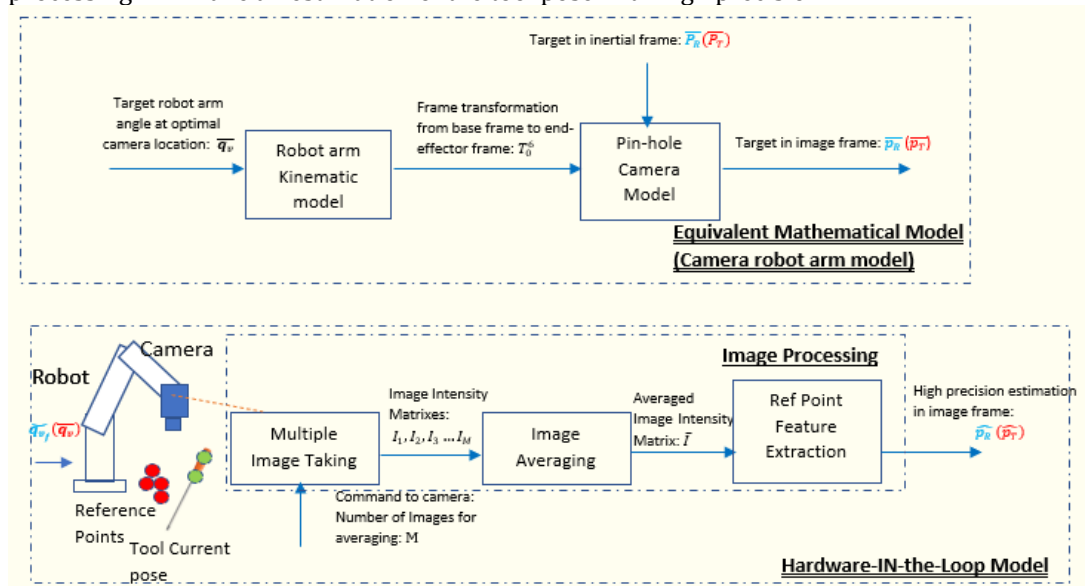

Figure 9.  The high precision camera sensor.
Note: Blue are signals used in the control system of Figure (7) and red are signals used in the control system of Figure (**8**).

## 5. The inner-loop control design

In this part, we will present the control system design of the camera pose control and the inner joint control loops for both the camera movement adjustment and the tool manipulation, which have been introduced in Section 4. A simulation scenario is also presented in this section.

We express an equation for the 6 DOF manipulator including the dynamics of the robot manipulator and the actuators (DC motor) in the following simplified form [16]:

$$(D(q) + J)\ddot{q} + (C(q,\dot{q}) + \frac{B}{r})\dot{q} + g(q) = u \tag{19}$$

where $D(q)$ and $C(q,\dot{q})$ are $6 \times 6$ inertial and Coriolis matrices respectively. $J$ is a diagonal matrix expressing the sum of actuator and gear inertias. $B$ is the damping factor, and $r$ is the gear ratio, $g(q)$ is the term for potential energy, $u$ is the 6 x 1 input vector, and $q$ is the 6 x 1 generalized coordinates (in this paper, $q$ is the 6 x 1 joint angle vector).

We can simplify Eq. (19) as follows:

$$M(q)\ddot{q} + h(q,\dot{q}) = u \tag{20}$$

with
$$M(q) = D(q) + J \tag{21}$$

$$h(q,\dot{q}) = (C(q,\dot{q}) + \frac{B}{r})\dot{q} + g(q) \tag{22}$$

If we transform the control input as following:
$$u = M(q)v + h(q,\dot{q}) \tag{23}$$

where $v$ is a virtual input. Then we substitute for u in Equation (20) using (23), and since $M(q)$ is invertible, we will have a reduced system equation as follows:

$$\ddot{q} = v \tag{24}$$

This transformation is so-called feedback linearization technique with the new system equation given in Eq. (24). This equation represents 6 uncoupled double integrators. The overall feedback linearization method is illustrated in Figure 10. In this control block diagram, we force the joint angle $q$ to follow the target joint angle $q_R$ so that the output pose, $P$, can follow the target pose, $\bar{P}$. $P$, $\bar{P}$, $q$, and $q_R$ are all vectors with six elements (each element corresponds to a joint position or angle). The Nonlinear interface transform the linear virtual control input $v$ to the nonlinear control input u by using Eq. (23). The output of the manipulator dynamic model, the joint angles, $q$, and their first derivatives, $\dot{q}$, are utilized to calculate $M(q)$ and $h(q,\dot{q})$ in the Nonlinear interface. The linear joint controller is designed using Youla parameterization technique [38] to control the nominally linear system in Eq. (24).

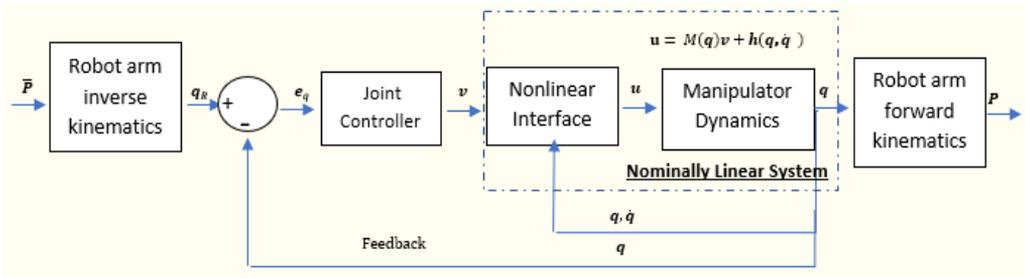

Figure 10. The block Diagram of feedback linearization Youla control design used for the joint control loop.

The design of a linear Youla controller with nominally linear plant is presented next.

Since the transfer functions between all inputs to outputs in (24) are the same and decoupled, we can first design a SISO (single input and single output) controller and use the multiple of the same controller for a six-dimension to obtain the MIMO (Multiple Input Multiple Output) version. In other words, we first design a controller $G_{c_{SISO}}$ that satisfies:

$$v_{SISO} = \ddot{q}_{SISO} \qquad (25)$$

where $v_{SISO}$ is a single input to a nominally linear system and $\ddot{q}_{SISO}$ is the second order derivative of a joint angle. The controller system in Figure 10 can be then written as:

$$G_{c_{sys}} = G_{c_{SISO}} \cdot I \qquad (26)$$

where $I$ is a 6 × 6 identity matrix. We can design the SISO controller $G_c$ using Youla parameterization technique [38]. The transfer function of the SISO nominally linear system from (24) is:

$$G_{p_{SISO}} = \frac{1}{s^2} \qquad (27)$$

Note that $G_{p_{SISO}}$ has two BIBO (Bounded Input Bounded Output) unstable poles at origin. To ensure internal stability of the feedback loop, the closed loop transfer function, $T_{SISO}$, should meet the interpolation conditions [38]:

$$T_{SISO}(s=0) = 1 \qquad (28)$$

$$\frac{dT_{SISO}}{ds}\bigg|_{s=0} = 0 \qquad (29)$$

We compute a Youla transfer function: $Y_{SISO}$, using the following relationship,

$$T_{SISO} = Y_{SISO} G_{p_{SISO}} \qquad (30)$$

The $T_{SISO}$ is designed so that it satisfies the conditions in (28) and (29). The sensitivity transfer function, $S_{SISO}$, is then calculated as follows:

$$S_{SISO} = 1 - T_{SISO} \qquad (31)$$

Without providing the design details, we refer the interested reader to [38], the closed-loop transfer function should be in the following form to satisfy the interpolation conditions:

$$T_{SISO} = \frac{(3\tau s + 1)}{(\tau s + 1)^3} \qquad (32)$$

Where $\tau$ specifies the pole and zero locations and represents the bandwidth of the control system. We can tune $\tau$ so that the response can be fast with less-overshoot.

Then we can derive $G_{c_{SISO}}$ from relationships between the closed-loop transfer function, $T_{SISO}$, the sensitivity transfer function, $S_{SISO}$, and the Youla transfer function, $Y_{SISO}$, in (30)-(32):

$$Y_{SISO} = T_{SISO} G_{p_{SISO}}^{-1} = \frac{s^2(3\tau^2 s + 1)}{(\tau s + 1)^3} \qquad (33)$$

$$S_{SISO} = 1 - T_{SISO} = \frac{s^2(\tau^3 s + 3\tau^2)}{(\tau s + 1)^3} \qquad (34)$$

$$G_{c_{SISO}} = Y_{SISO} S_{SISO}^{-1} = \frac{3\tau^2 s + 1}{\tau^3 s + 3\tau^2} \qquad (35)$$

From Eq (35), we can compute a MIMO controller as follows:

$$G_{c_{sys}} = \frac{3\tau^2 s + 1}{\tau^3 s + 3\tau^2} \cdot I_{6\times 6} \qquad (36)$$

Equation (36) provides the desired controller, which is used as the joint controller, as shown in Figure 10. This configuration is precisely the inner joint control loop in both the visual and the manipulator systems as shown in Figure 7 and 8.

Figure 11 shows the simulation results for the case with no disturbance. The target position and the orientation of the end-effector are selected to be $\begin{bmatrix}\bar{X}\\\bar{Y}\\\bar{Z}\end{bmatrix} = \begin{bmatrix}1.7157m\\1.0191m\\0.7518m\end{bmatrix}$ and $\begin{bmatrix}\bar{n}\\\bar{s}\\\bar{a}\end{bmatrix} = \begin{bmatrix}-0.425 & 0.87 & 0.25\\0.8361 & 0.2714 & 0.4767\\0.3469 & 0.4116 & -0.8428\end{bmatrix}$, where $[\bar{X},\bar{Y},\bar{Z}]^T$ is the absolute position coordinate of the center of the end effector in the inertial frame and $\bar{n},\bar{s},\bar{a}$ represent respectively the end-effector's directional unit vector of the yaw, pitch and roll in the inertial frame. Therefore, the corresponding target angles of rotations are $q_R = [30°, 60°, -45°, 15°, 45°, 90°]$. For this simulation, we have designed the control system with the bandwidth of $100\ rad/s$. In the following three plots, solid lines represent the responses for the end-effector position of each joint and the end-effector orientation respectively, and the dashed lines are the targets. Specifically, the orientation response of the end-effector is the vector that tangent to the curve in the second plot at each point in Figure 11. The simulation results show that all responses of the controlled system will be able to reach their final/steady state values within 0.1 second with no (or little) overshoots.

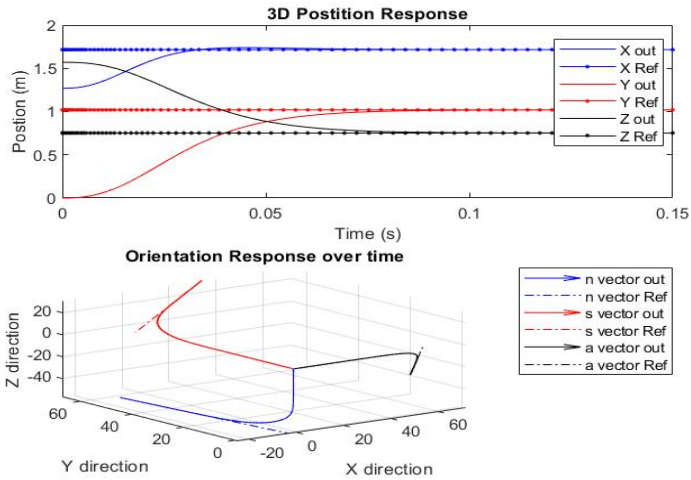

Figure 11. The simulation results for the end-effector response to an arbitrary selected trajectory.

## 6 The cascaded SISO outer-loop design for the camera movement adjustment control system

As introduced in Section 4.1, we use the Image Based Visual Stereo (IBVS) as the framework for cascaded control design of both the visual and the manipulation system. The inner-loop is the joint control loop, as discussed previously. In the IBVS, an outer feedback control is designed in addition to the inner feedback control so that the 2D visual features can be compared and matched. Therefore, a camera

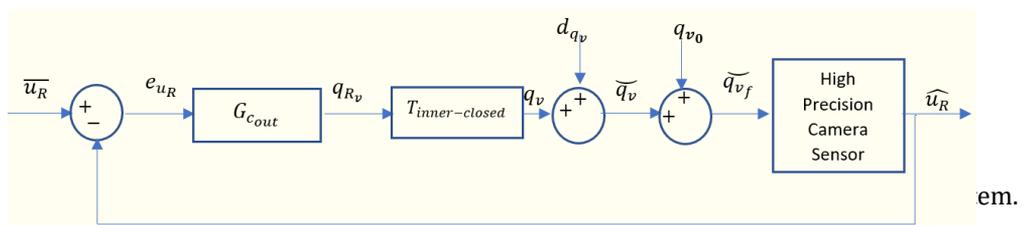

em.

model is required in the feedback loop to map the joint angles to the object visual features (e.g., 2D coordinates of the object in the image frame). To start from a simple case, we discuss the requirements for the design of the cascaded control for a SISO system. Assume, we only measure and try to control one feature: the coordinate of the object in one axis, then the overall cascaded control diagram of this configuration is shown in Figure 12.

$\overline{u_R}$ is the target of $u$ coordinate of a point (one reference point in visual system). $G_{c_{out}}$ is the outer-loop controller, which provides the target joint angle $q_{R_v}$ to the robot arm based on the image coordinate error $e_{u_R}$. A combined uncertainty signal, $d_{q_v}$ (e.g., sensor noise, backlash, friction, and compliance due to gear reduction in the joint) is added to the joint angle of rotation $q_v$, the output of the inner control loop. Furthermore, the initial joint angle of robot manipulator, $q_{v_0}$, is added to the output joint angle with disturbance, $\widetilde{q_v}$, to generate the final joint angle $\widetilde{q_{v_f}}$.

### 6.1 The camera sensor model for the SISO control design

In this section, we are going to present a simplified mathematical model of the camera. Let us consider a one-link robot manipulator with a camera that can be rotated around the $Z$-axis, as shown in Figure 13. Let us also consider a point that is located in the $X - Y$ plane.

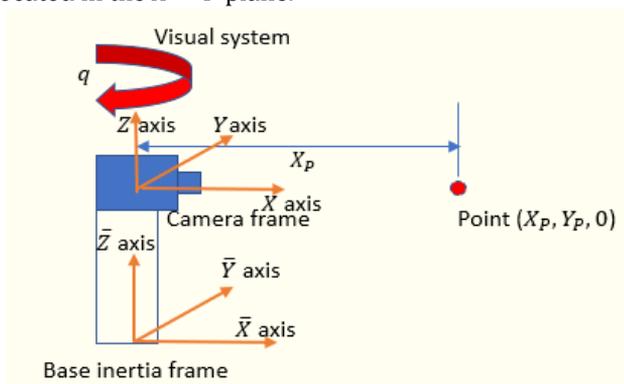

Figure 13. One-link manipulator with a camera.

We use a Cartesian coordinate frame $OXYZ$, which is attached at the camera center. The $X$-axis is perpendicular to the camera lens and the coordinate frame rotates around $Z$-axis. A point whose coordinate is $(X_P, Y_P, 0)$ in the rotational frame is projected on the image $u - v$ plane with its coordinates as $(u, 0)$. In addition, an inertial frame $\overline{O}\overline{X}\overline{Y}\overline{Z}$ is shown at the base of the manipulator in Figure 13.

We derive the equations for a pin-hole camera model (a monocular case in stereo model in Section 2.1). Assuming the skew coefficient and image coordinate offsets are zero ($s_c = u_0 = v_0 = 0$), then from Eq. (1), the one-dimensional image coordinate u can be written as:

$$u = \frac{f_u Y_P}{X_P} \qquad (37)$$

The coordinate system, located at the camera center, rotates with the camera. Assume the camera starts at the position where the $X$-axis is parallel to the $\bar{X}$-axis of the inertial coordinate system and positive angle is defined when the camera rotates in a clockwise direction with an angle $q$ around $Z$-axis. After the frame rotates $q$

clockwise, the new coordinates of the point in the new coordinate frame $(X'_P, Y'_P, 0)$ can be computed as:

$$\begin{bmatrix} X'_P \\ Y'_P \end{bmatrix} = \begin{bmatrix} \cos(q) & -\sin(q) \\ \sin(q) & \cos(q) \end{bmatrix} \begin{bmatrix} X_P \\ Y_P \end{bmatrix} \quad (38)$$

Combining Eqs. (37) and (38), the new image coordinate, $u'$, can be calculated as:

$$u' = \frac{f_u Y'_P}{X'_P} = f_u \frac{X_P \sin(q) + Y_P \cos(q)}{X_P \cos(q) - Y_P \sin(q)} \quad (39)$$

Let us write the coordinates $(X_P, Y_P)$ in the polar coordinates, such that,

$$R = \sqrt{X_P^2 + Y_P^2}$$

and 
$$X_P = R\cos(\varphi), Y_P = R\sin(\varphi),$$

where 
$$\varphi = \tan^{-1}(Y_P/X_P) \quad (40)$$

Therefore, Eq. (39) can be written in the polar form as:

$$u' = f_u \frac{R\sin(\varphi + q)}{R\cos(\varphi + q)} = f_u \tan(\varphi + q) \quad (41)$$

The angle $\varphi$ is the angle of the point with respect to the $\bar{X}$-axis and the angle $q$ is is already defined. Eq. (41) mathematically expresses a model of the camera sensor shown in the block diagram of Figure 13.

### 6.2 Outer-loop controller design

In the previous sections, we presented the results for designing inner joint loop controllers using Youla parameterization method. In the following two sections, we will discuss the design of the outer-loop controllers by using two different approaches:

a) Feedback linearization
b) Model linearization

### 6.2.a Feedback linearization

Let us relook at the cascaded block diagram in Figure 12. The inner closed loop transfer function is already derived in (32):

$$T_{inner-closed} = \frac{q_v(s)}{q_{R_v}(s)} = \frac{(3\tau_{in}s + 1)}{(\tau_{in}s + 1)^3} \quad (42)$$

Notice we replaced $\tau$ with $\tau_{in}$ to indicate bandwidth of inner-closed loop. Using equation (41) and considering the block diagram in Figure 12, we can write:

$$\widehat{u_R} = f_u \tan\left(\varphi + \widetilde{q_{v_f}}\right), \text{with } \varphi = \tan^{-1}(Y_P/X_P) \quad (43)$$

We can rewrite (42) and (43) in time domain, by introducing an intermediate variable or state, $W$, as:

$$\tau_{in}\dddot{W} + 3\tau_{in}^2\ddot{W} + 3\tau_{in}\dot{W} + W = q_{R_v} \tag{44}$$

$$f_u tan(\varphi + q_{v_0} + 3\tau_{in}\dot{W} + W) = \widehat{u_R} \tag{45}$$

Eqs. (44) and (45) describe a nonlinear third order system, where $W$ is the state, $q_{R_v}$ is the input and $\widehat{u_R}$ is the output. We can use feedback linearization method to design the outer-loop controller by taking the second order derivative of (45) and combine with (44) to obtain:

$$\ddot{\widehat{u_R}} = R(W,\dot{W},\ddot{W}) + G(W,\dot{W},\ddot{W})q_{R_v} \tag{46}$$

where

$$\begin{aligned}R(W,\dot{W},\ddot{W}) &= 2f_u cos^{-2}(\varphi + q_{v_0} + 3\tau_{in}\dot{W} + W)\tan(\varphi + q_{v_0} \\ &+ 3\tau_{in}\dot{W} + W)(3\tau_{in}\ddot{W} + \dot{W})^2 \\ &-f_u cos^{-2}(\varphi + q_{v_0} + 3\tau_{in}\dot{W} + W)\left(8\ddot{W} + \frac{9}{\tau_{in}}\dot{W} + \frac{3}{\tau_{in}^2}W\right)\end{aligned} \tag{47}$$

$$G(W,\dot{W},\ddot{W}) = f_u cos^{-2}(\varphi + q_{v_0} + 3\tau_{in}\dot{W} + W)\frac{3}{\tau_{in}^2} \tag{48}$$

We can transform or map these variables so that the nonlinear system in (47)-(48) can be written as an equivalent linear state-space representation as follows:

$$\varepsilon_1 = f_u tan(\varphi + q_{v_0} + 3\tau_{in}\dot{W} + W), \tag{49}$$

$$\varepsilon_2 = \dot{\varepsilon}_1 = f_u cos^{-2}(\varphi + q_{v_0} + 3\tau_{in}\dot{W} + W) \cdot (3\tau_{in}\ddot{W} + \dot{W}) \tag{50}$$

We can write the state-space form of (49) and (50) as:

$$\dot{\varepsilon} = \begin{bmatrix}\dot{\varepsilon}_1\\\dot{\varepsilon}_2\end{bmatrix} = \begin{bmatrix}0 & 1\\0 & 0\end{bmatrix}\begin{bmatrix}\varepsilon_1\\\varepsilon_2\end{bmatrix} + \begin{bmatrix}0\\1\end{bmatrix}U \tag{51}$$

$$\widehat{u_R} = [1\ 0]\begin{bmatrix}\varepsilon_1\\\varepsilon_2\end{bmatrix} \tag{52}$$

where 
$$U = G(W,\dot{W},\ddot{W})q_{R_v} + R(W,\dot{W},\ddot{W}) \tag{53}$$

Transform the state-space representation back to the transfer function form, we can write:

$$Gp_{nominal} = \frac{\widehat{u_R}(s)}{U(s)} = C(sI - A)^{-1}B = \frac{1}{s^2} \tag{54}$$

where 
$$A = \begin{bmatrix}0 & 1\\0 & 0\end{bmatrix}, B = \begin{bmatrix}0\\1\end{bmatrix}, and\ C = [1\ 0] \tag{55}$$

Since the $Gp_{nominal}$ is the same as the plant transfer function in (27), the design of Youla controller for this linear system is similar to (28) to (35). Therefore, the transfer function of the outer-loop controller can be written as:

$$G_{C_{out}} = \frac{3\tau_{out}^2 s + 1}{\tau_{out}^3 s + 3\tau_{out}^2} \tag{56}$$

$\tau_{out}$ determines the pole and zero locations of closed-loop transfer function of the outer-loop controller and therefore, represents the bandwidth of outer-loop controller. We must make sure that $\tau_{out} > \tau_{in}$ so that inner-loop responds faster than the outer-loop in the cascaded control design strategy. The overall block diagram is shown in Figure 14.

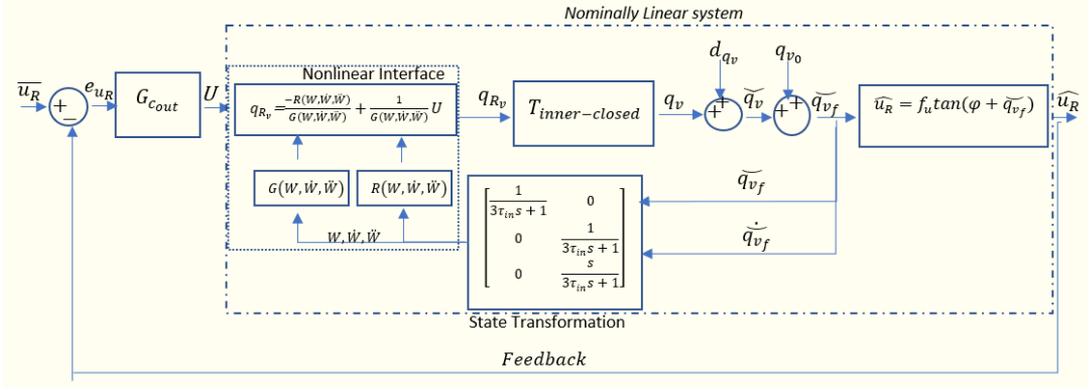

Figure 14. The block Diagram of the SISO outer-loop control with feedback linearization for robot arm movement adjustment.

The states $W, \dot{W}, \ddot{W}$ are computed from the rotational angle and its derivative $\widetilde{q_{v_f}}$ and $\dot{\widetilde{q_{v_f}}}$. From Eq. (42), we can obtain the following relationship:

$$\widetilde{q_{v_f}} = 3\tau_{in}\dot{W} + W \tag{57}$$

Therefore, the transfer function from $W$ to $\widetilde{q_{v_f}}$ can be written as:

$$W(s) = \frac{1}{3\tau_{in}s + 1}\widetilde{q_{v_f}}(s) \tag{58}$$

Also, we can obtain the transfer function of $\dot{W}(s)$ and $\ddot{W}(s)$ as:

$$\dot{W}(s) = \frac{1}{3\tau_{in}s + 1}\dot{\widetilde{q_{v_f}}}(s) \tag{59}$$

$$\ddot{W}(s) = \frac{s}{3\tau_{in}s + 1}\dot{\widetilde{q_{v_f}}}(s) \tag{60}$$

Eqs (58) to (60) are shown as the state transformation block in Figure 14.

It is worthwhile to note that the equation in the nonlinear interface:

$$q_{R_v} = \frac{-R(W,\dot{W},\ddot{W})}{G(W,\dot{W},\ddot{W})} + \frac{1}{G(W,\dot{W},\ddot{W})}U \tag{61}$$

is only defined when $G(W, \dot{W}, \ddot{W}) = f_u \cos^{-2}(\varphi + q_{v_0} + q_v)\frac{3}{\tau_{in}^2} \neq 0$. It can be shown that this is always not equal to zero and $\varphi + q_{v_0} + q_v \neq \pm\frac{\pi}{2}$. This is always true because any angle of rotation should be within half of camera's angle of view $\alpha$; that is $\left|\varphi + \widetilde{q_{v_f}}\right| \leq \frac{\alpha}{2} < \frac{\pi}{2}$. In addition, $\alpha$ is always less than $\pi$ for any camera type. Therefore, the controller works for the entire range independent of the camera type.

### 6.3 Model linearization

We can deal with the nonlinear system by linearizing the system first and then design a linear controller using the system transfer function. In Figure 12, $T_{inner-closed}$ transfer function is given in (42) and the nonlinear form of the camera model is provided in (43). The overall dynamic system combines the inner-loop and the camera model, which will be linearized so that the combined dynamic system will then be linear. Next, we linearize, the camera model, (43), around an equilibrium point $\widetilde{q_{v_0}}$:

$$\widehat{u_R} = f_u\cos^{-2}(\varphi + \widetilde{q_{v_0}})(\widetilde{q_{v_f}} - \widetilde{q_{v_0}}) + f_u\tan(\varphi + \widetilde{q_{v_0}}), \tag{62}$$

If we assume $\widetilde{q}_{v_0} = 0$, then:
$$\widehat{u_R} = f_u \cos^{-2}(\varphi)\widetilde{q}_{v_f} + f_u \tan(\varphi) \quad (63)$$

Assuming $C_1 = f_u \cos^{-2}(\varphi)$, $C_2 = f_u \tan(\varphi)$, therefore, Eq. (63) can be rewritten as:
$$\widehat{u_R} = C_1 \widetilde{q}_{v_f} + C_2 \quad (64)$$

Let us define $\widehat{u_R}' = \widehat{u_R} - C_2$, then, the overall block diagram of the linearized system is shown in Figure 15.

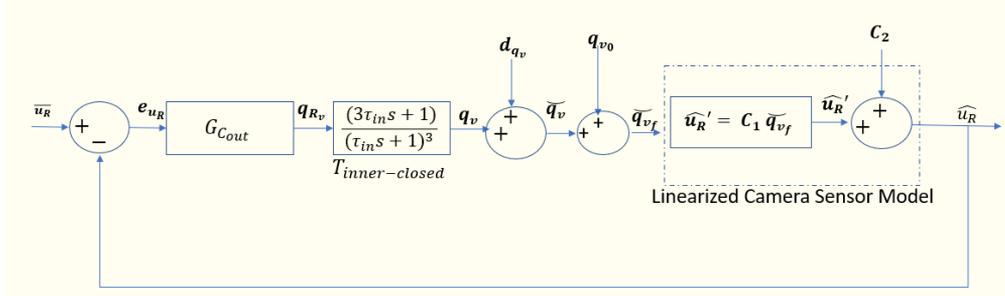

Figure 15. The block Diagram of the SISO control with the linearized camera model for robot arm movement adjustment.

.

The plant transfer function is derived as:
$$G_{p_{linear}} = \frac{\widehat{u_R}'}{q_{R_v}} = C_1 \frac{(3\tau_{in}s + 1)}{(\tau_{in}s + 1)^3} \quad (65)$$

The design of a Youla controller is trivial in this case as all poles/zeros of the plant transfer function in (65) are located in the left half-plane, and therefore, they are stable. In this case, we can shape the closed loop transfer function, $T_{out}$, by selecting a Youla transfer function: $Y_{out}$ so that the closed loop transfer function, $T_{out}$, will not contain any plant poles and zeros. All poles and zeros in the original plant can be cancelled out and new poles and zeros can be added to shape the closed-loop system. Let us select a Youla transfer function so that the closed-loop system behaves like a second order Butterworth filter, such that:

$$Y_{out} = \frac{1}{G_{p_{linear}}} \frac{\omega_n^2}{(s^2 + 2\zeta\omega_n s + \omega_n^2)} \quad (66)$$

then:
$$T_{out} = \frac{\omega_n^2}{(s^2 + 2\zeta\omega_n s + \omega_n^2)} \quad (67)$$

where $\omega_n$ is called natural frequency and approximately sets the bandwidth of the closed–loop system. We need to make sure the bandwidth of the outer-loop is smaller than the inner-loop, i.e., $1/\omega_n > \tau_{in}$. $\zeta$ is called the damping ratio, which is another tuning parameter.

Using Eqs. (34) and (35), we can calculate the sensitivity transfer function, $S_{out}$, and the controller transfer function, $G_{C_{out}}$, of the outer-loop in cascaded control design as:

$$S_{out} = 1 - T_{out} = \frac{s^2 + 2\zeta\omega_n s}{(s^2 + 2\zeta\omega_n s + \omega_n^2)} \quad (68)$$

$$G_{C_{out}} = Y_{out} S_{out}^{-1} = \frac{1}{C_1} \frac{(\tau_{in}s + 1)^3}{(3\tau_{in}s + 1)} \frac{\omega_n^2}{(s^2 + 2\zeta\omega_n s)} \quad (69)$$

## 6.4 The simulation of the cascaded SISO closed-loop system for camera movement adjustment

In this section, we are going to compare the closed-loop response results of the cascaded control system where the outer-loop controllers are designed using the two aforementioned methods: feedback linearization (Section 6.2) and model linearization (Section 6.3). The simulation results are obtained with the original nonlinear camera model (43). In addition, for the linearized plant approach, we will also illustrate how varying the damping ratio ζ affects the responses. For both methods, we chose the bandwidth of the inner-loop as $100 rad/s$ and the bandwidth of the outer-loop as $10 rad/s$.

We compare the simulation responses by choosing six different damping ratios. Two are chosen as the overdamped systems (ζ>1), one is chosen as a critically damped system (ζ=1), and three as the underdamped systems (ζ<1). We have simulated four cases and compared all six systems for each case. Each case is different due to varying the initial angle $\varphi$ (see Eq. (40)) and the input disturbance $d_{q_v}$. Two cases are simulated without the input disturbance while the other two are simulated with the disturbance to compare the robustness of the controlled system.

The step responses of the image coordinate $\widehat{u_R}$ is shown in Figures 16, 17, 18, and 19. The intrinsic camera parameters are selected to be: $f_u = 2.8\ mm$ (Focal length) and $\alpha = 120°$ (Angle of view). In cases 1 and 2, the responses of feedback linearization are displayed in black dashed lines while all other lines are the responses of the controller that is designed with the linearized plant and varying the damping ratio ζ.

The case 1 is simulated with the initial angle $\varphi < \frac{\alpha}{2}$, while the case 2 is simulated when $\varphi = \frac{\alpha}{2}$, the largest possible initial angle within the angle of view. It can be shown clearly that without any input disturbance, both methods are able to drive the closed-loop responses to the final value. The step response of the feedback linearization has an overshoot. In addition, for the second method, the model linearization approach, it can be seen from the two simulation cases that there exists a damping ratio, $\zeta_{opt}$, such that

- When $\zeta \geq \zeta_{opt}$, the step responses have no overshoots and as ζ decreases, the system reaches the steady state faster.
- When $\zeta < \zeta_{opt}$, the step responses have overshoots, and the overshoots increase as ζ decreases. As ζ increases, the system reaches the steady state faster.

It can be estimated from Figures 16 and 17 that $\zeta_{opt} \cong 1$ in the case 1 and $\zeta_{opt} \cong 0.5$ in the case 2. The most desirable system is the one without overshoot and fastest step response. When $\zeta = \zeta_{opt}$, the system has the fastest response and no (or little) overshoot. Therefore, we can state that the best performance of the controlled system is when setting the damping ratio $\zeta = \zeta_{opt}$. Clearly, the value of $\zeta_{opt}$ varies with $\varphi$, the initial angle of the reference point with respect to the inertial frame.

In the cases 3 and 4, the input disturbance is introduced to the system. In the case 3, a small disturbance ($d_{q_v} = 5°$) is added to the actuator input. The case 4 is a combined case where both $\varphi$ and $d_{q_v}$ are present ($\varphi = 20°, d_{q_v} = -60°$). Figures 18 and 19 don't display the step responses of feedback linearization approach. The step responses of feedback linearization are unstable when input disturbances are introduced. It can be shown that any input disturbance drastically alters the nonlinear interface parameters, used in feedback linearization, and hence, results in an unstable system. On the other hand, the linear controller designed based on the linearized plant model is robust to the input disturbances even with the significantly large disturbances (case 4). Similar to the no disturbance cases, $\zeta_{opt}$ exists for the cases with disturbances.

From the discussion above, the plant linearization method is the preferred and the recommended method for the camera movement adjustment.

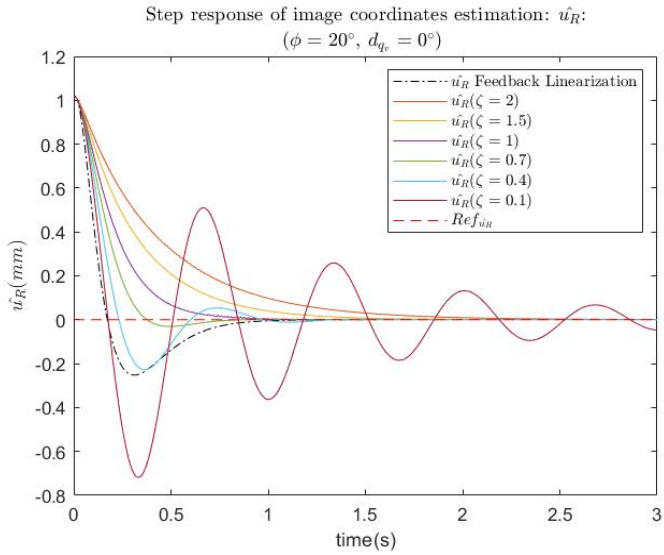

Figure 16. Step responses of $\widehat{u_R}$ for the case 1: $\varphi < \frac{\alpha}{2} = 20°$, $d_{q_v} = 0°$.

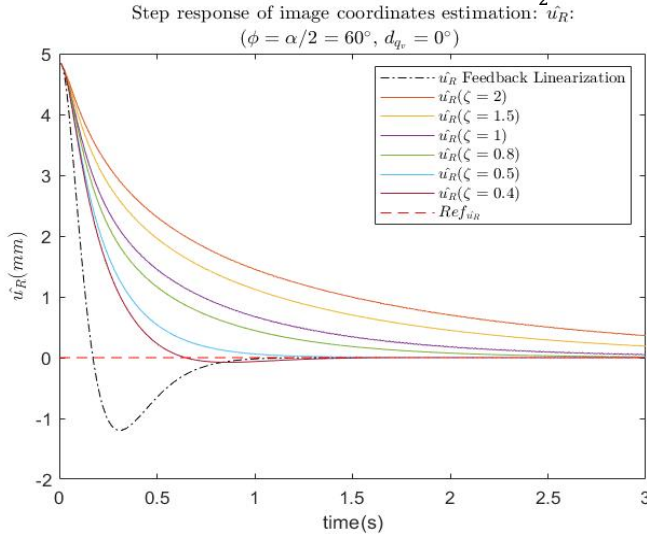

Figure 17. Step responses of $\widehat{u_R}$ for the case 2: $\varphi = \frac{\alpha}{2} = 60°$, $d_{q_v} = 0°$.

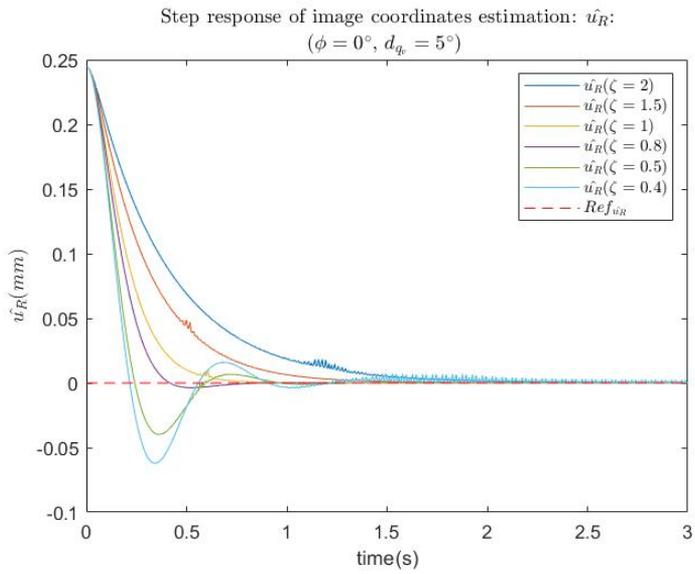

Figure 18. Step responses of $\widehat{u_R}$ for the case 3: $\varphi = 0°$, $d_{q_v} = 5°$.

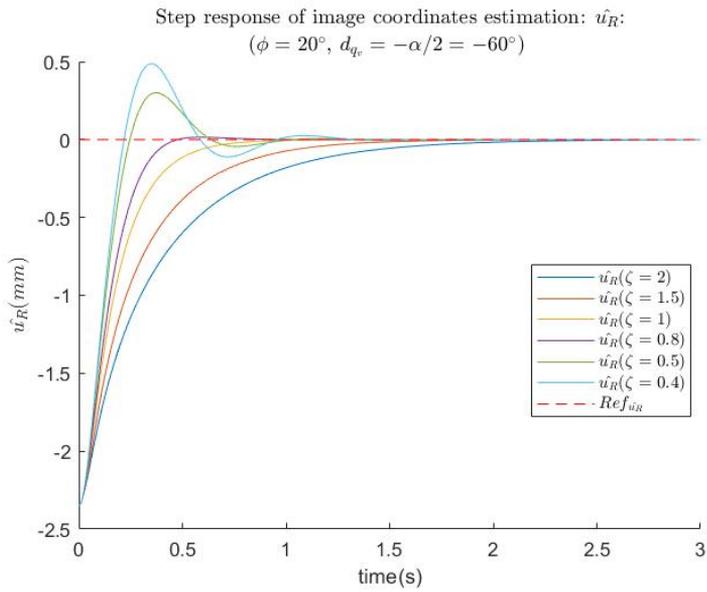

Figure 19. Step responses of $\widehat{u_R}$ for the case 4: $\varphi = 20°$, $d_{q_v} = -\frac{\alpha}{2} = -60°$.

# 7 The cascaded SISO outer-loop design for high-accuracy robot tool manipulator

The control block diagram of high-accuracy robot tool manipulator is shown in Figure 8. In this diagram, the camera is kept static but servos the movement of the tool by the visual data. This block diagram contains a feedback loop as well as a feedforward loop. We design controllers for each loop and simulate the combined loop under different scenarios.

## 7.1 Developing a combined SISO tool robot arm and camera model

In Figure 8, the joint angle including the disturbance of the tool manipulator is transformed to the tool pose using a robot kinematics model. A camera model then is utilized to convert the 3D pose to the 2D, as shown in Figure 20. For simplicity, we can combine these two blocks into one block, which is called the tool robot arm and camera sensor model.

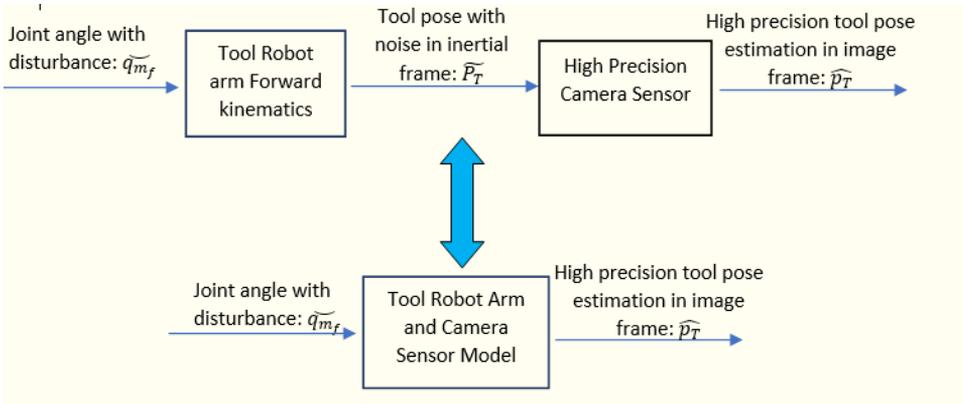

Figure 20. The tool robot arm and the camera sensor model (a combined block of tool manipulation kinematics and camera sensing).

In Figure 21, a SISO combined model setup has been shown based on the one-link camera robot arm model in Figure 13. Now, the camera, which is attached to a one-link rotational robot arm captures the image of a tool, which is attached to another similar robot arm, and estimates its angle of rotation. The tool has a length $L_t$ with an interest point is selected at the tip of the tool. Both robot links have a length of $L_1$ and are separated from each other by a distance $L$. Assume $\overline{q_v}$ is the angle of the camera from previous control sequences (discussed in Figure 5). The inertial and camera coordinate frames set ups are discussed in Section 6.1. The only difference is that the camera frame rotates relative to inertial frame by a clockwise angle $\overline{q_v}$ along the $\bar{Z}$-axis. The tool rotates relative to the $\bar{Z}$-axis in a clockwise direction with a variable angle $\widetilde{q_m}$. The actual angle of rotation $\widetilde{q_m}$ is the sum of the input disturbance $d_{q_m}$ and the planned angle of rotation $q_m$; i.e.,

$$\widetilde{q_m} = q_m + d_{q_m} \quad (70)$$

Then, we can compute the final angle after rotation by adding the initial angle of the tool in the inertial frame $q_{m_0}$:

$$\widetilde{q_{m_f}} = \widetilde{q_m} + q_{m_0} \quad (71)$$

The coordinates of the point of interest on the tool in the inertial frame is then computed as $(L - L_t \cos(\widetilde{q_{m_f}}), L_t \sin(\widetilde{q_{m_f}}), L_1)$.

Following the same procedures as in Eqs (37) to (41), we can derive the tool image coordinate $\widehat{u_T}$ along the $u$ -axis as:

$$\widehat{u_T} = f_u \frac{Q(\widetilde{q_{m_f}}) + \tan(\overline{q_v})}{1 - Q(\widetilde{q_{m_f}})\tan(\overline{q_v})} \tag{72}$$

where,
$$Q(\widetilde{q_{m_f}}) = \frac{L_t \sin(\widetilde{q_{m_f}})}{L - L_t \cos(\widetilde{q_{m_f}})} \tag{73}$$

Eqs. (72) and (73) provide a function that maps the current or the final angle of the tool onto the image coordinate $\widehat{u_T}$ with constant parameters, $\overline{q_v}$, $L$, and $L_t$.

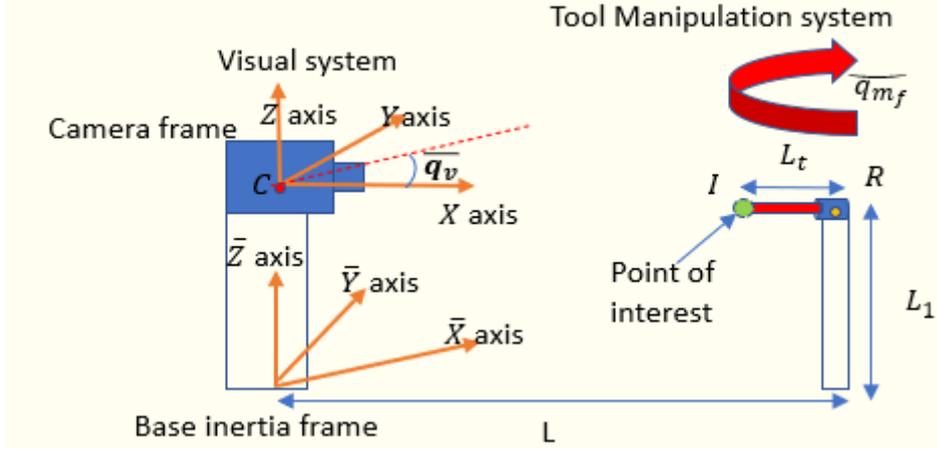

Figure 21. A SISO camera and tool robot arm setup.

### 7.2 The outer-loop feedback and feedforward controller design

The overall plant for the design of this control system is composed of the inner joint control loop, see Eq. (32) and Figure 8, and the tool robot arm and camera sensor models, as shown in Figures 20. We can design the outer-loop feedback controller using the feedback linearization method or the plant linearization method by following the procedures presented in Section 6.2 and 6.3 respectively. For the sake of brevity, we won't discuss the detail derivations of each controller. Mostly comparative issues such the overshoots and the robustness are discussed in Section 6.4. In this section, we utilize the plant linearization method to design the outer-loop feedback controller.

Without providing the details, the controller is designed for a second order closed -loop system using the plant linearization method (the plant is linearized at $\widetilde{q_{m_f}} = 0°$) is given as:

$$G_{c_{out}} = \frac{1}{C_1} \frac{(\tau_{in} s + 1)^3}{(3\tau_{in} s + 1)} \frac{\omega_n^2}{s^2 + 2\zeta \omega_n s} \tag{74}$$

and
$$C_1 = f_u (1 + (\tan(\overline{q_v}))^2 \frac{LL_t - L_t^2}{(L - L_t)^2} \tag{75}$$

Then, the second order closed-loop transfer function $T$ of the overall cascaded control system is expressed as:

$$T = \frac{\omega_n^2}{s^2 + 2\zeta \omega_n s + \omega_n^2} \tag{76}$$

Where $f_u$ is the camera focal length, $\overline{q_v}$, $L$, and $L_t$ are the parameters defined in Section 7.1. $\tau_{in}$ defines the bandwidth of the inner joint loop. $\omega_n$ is the natural frequency and $\zeta$ is the damping ratio of the second order system.

As the camera is static in this control stage, the tool pose cannot be recognized and measured visually if it is outside the camera range of view. To tackle this problem, we can estimate the 2D feature (image coordinates of the tool points) from the same model in Eqs. (72) and (73) with the joint angle $q_m$ as input:

$$\widetilde{u_T} = f_u \frac{Q(q_m) + \tan(\overline{q_v})}{1 - Q(q_m)\tan(\overline{q_v})} \quad (77)$$

where
$$Q(q_m) = \frac{L_t \sin(q_m)}{L - L_t \cos(q_m)} \quad (78)$$

which is illustrated in the block diagram of Figure 22. Normal feedback loop (in blue lines) is preserved when the tool is inside the camera range of view and hence, the camera can estimate the tool 2D feature $\widehat{u_T}$. However, when the tool is outside the range of view, the 2D feature can only be approximated as $\widetilde{u_T}$ (red dashed line) by the combined model as shown in the blue dashed box. We can implement a bump-less switch to smoothly switch between these modes of operations. The switching signal changes over when the tool moves in or out of the camera range of view.

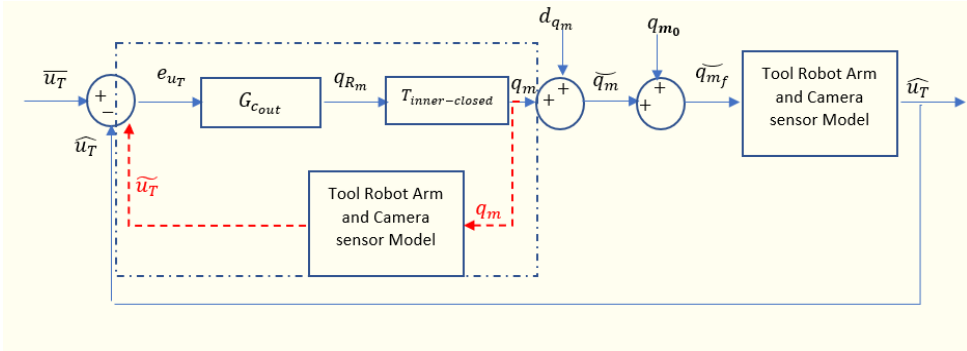

Figure 22. The block diagram of the tool manipulator feedback control loop with feature estimation.

In addition, the feedforward controller, as shown in Figure 8, is designed with the inverse kinematics model of the tool robot arm and the details of this design is not provided here. It should be noted, as stated previously, that the combination of the feedforward and the feedback controllers provide a much faster response than the feedback controller by itself.

## 7.3 The simulation of cascaded SISO closed loop for high-accuracy tool manipulation control system

In this section, we present the simulation results of the feedforward and the feedback control system designed for the robot tool manipulator. As discussed, the plant linearization method is used in the feedback controller design. The controlled system is simulated with the original nonlinear tool robot arm and the camera sensor model of Eqs. (77) and (78). Furthermore, we illustrate how varying the damping ratio $\zeta$ affects the responses. We chose the bandwidth of the inner-loop control as $100\,rad/s$ and the bandwidth of the outer-loop control as $10\,rad/s$. The intrinsic camera parameters are chosen as following: $f_u = 2.8\,mm$ (Focal length) and $\alpha = 120°$ (Angle of view). Other parameters are chosen as: $L = 1m$, and $L_t = 0.135m$.

We compare the performance of the feedforward-feedback control system and the feedback-only control system in two different scenarios. One scenario is simulated when the tool is kept inside the camera angle of view during the entire run time. It should be noted that both the feedback and the feedforward controllers are active during the entire simulation. The second scenario is simulated when the tool is outside the camera angle of view during the entire simulation time. When the tool pose is outside the camera angle of view, the feedback signal is replaced by the estimation signal from the model, as shown in Figure 22. Furthermore, for each scenario, we show how varying the damping ratio ζ affects the responses. Assuming the initial angle of the tool, $q_{m_0} = 0$ in the inertial base frame, we vary the pose or the rotational angle of the camera, $\overline{q_v}$, in the inertial base frame, for each simulation scenario.

Figures 23-26 show step responses of the joint angle $\widetilde{q_{m_f}}$ and the output tool pose $P_{T_X}$ (only X coordinate of the six dofs pose of the tool $P_T$) for the two scenarios. Each response is simulated with four different damping ratios: ζ=2 (Blue), ζ=1.5 (Green), ζ=1 (Purple), and ζ=0.7 (Black). With the same damping ratio (same color), the response of the feedback-only control system is shown in the dashed line and the response of the feedforward-feedback control system is illustrated in the solid line.

In the first scenario, the camera rotates 5° counterclockwise with respect to the inertial frame. Then, the tool stays in the camera angle of view with any joint angle $\widetilde{q_{m_f}} \in [-180°, 180°]$. In addition, a disturbance $d_{q_m} = -10°$, is added to this joint angle. The responses in Figures 23 and 24 illustrate that both the feedback-only and the feedforward-feedback control systems can reach stability and are robust to the disturbances. With varying the damping ratios, the feedforward-feedback system responses are faster in transient when compared to the response of the feedback-only system. The feedforward-feedback system response, when the damping ratio is small (ζ=0.7), results in a less overshoot when compared to the response of the feedback-only system. The optimal damping ratio, $\zeta_{opt} = 1$, (as discussed in Section 6.4), results in the fastest response and no overshoots, both for the feedforward-feedback and the feedback-only control systems.

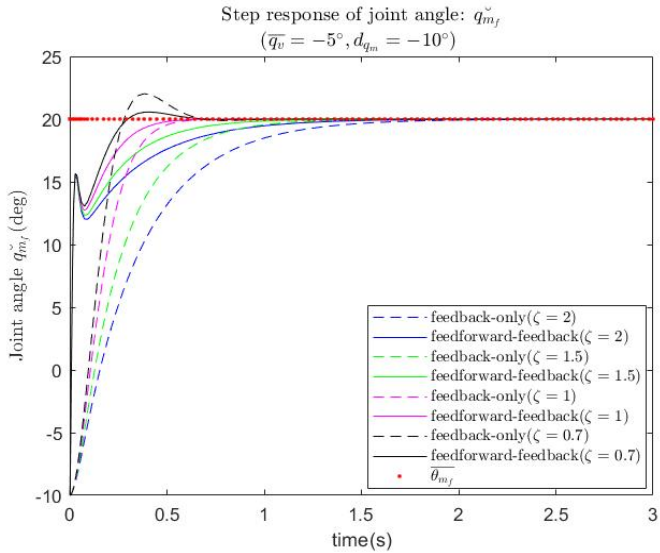

Figure 23. Step responses of $\widetilde{q_{m_f}}$. Scenario 1: $\overline{q_v} = -5°$, $d_{q_m} = -10°$.

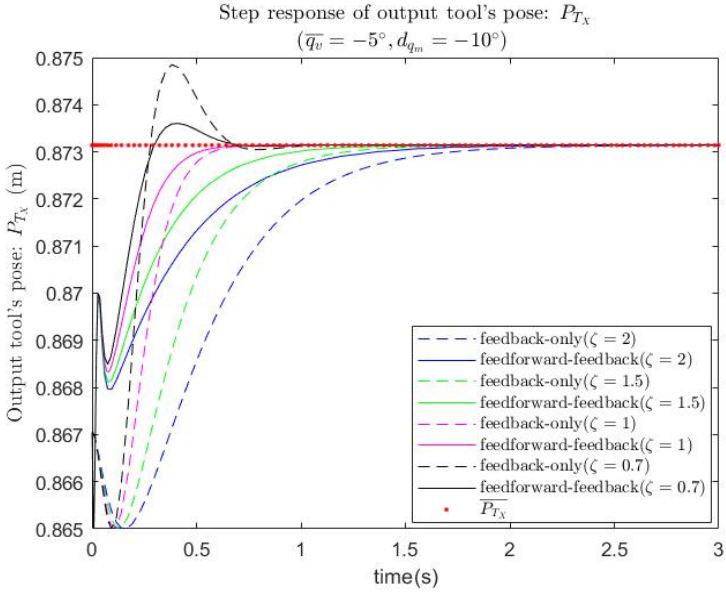

Figure 24. Step responses of $P_{T_X}$. Scenario 1: $\overline{q_v} = -5°$, $d_{q_m} = -10°$.

In the second scenario, the camera rotates 65° counterclockwise with respect to the inertial frame. Using geometry, we can calculate that the tool is out of the camera range of view when $\widetilde{q_{m_f}} \notin [35.21°, 134.79°]$. Staring from the initial angle $q_{m_0} = 0°$ and move to the target joint angle $\overline{q_{m_f}} = 50°$, there is a range $\widetilde{q_{m_f}} \epsilon [0°, 35.21°]$ that camera cannot detect the tool but estimation of the pose is required to drive the tool to the target. Even perturbed with the disturbance $d_{q_m} = 15°$, all the simulation responses shown in Figures (25) and (26) reach the target within a second. In this scenario, the feedforward-feedback control system still converges faster in transient but generates bigger overshoots compared to the feedback-only control system. The large overshoots may come from accumulated disturbances that cannot be eliminated by the feedforward control without the intervention of the feedback control. A feedforward controller may drive the tool away from its target even faster when the disturbance appears in the loop. Perhaps, a possible solution, which will be investigated in the future, would be the use of a switching algorithm, switching from a feedforward to a feedback controller, rather than the use of a continuous feedforward-feedback controller.

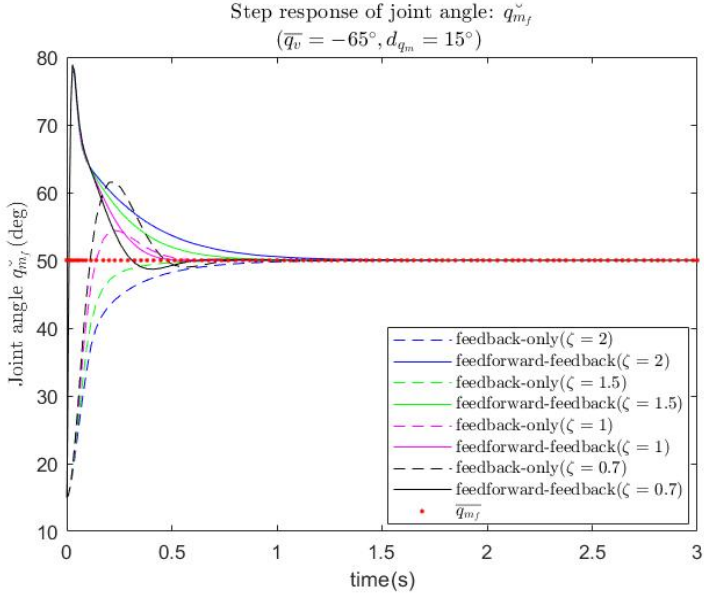

Figure 25. Step responses of $\widetilde{q_{m_f}}$. Scenario 2: $\overline{q_v} = -65°$, $d_{q_m} = 15°$.

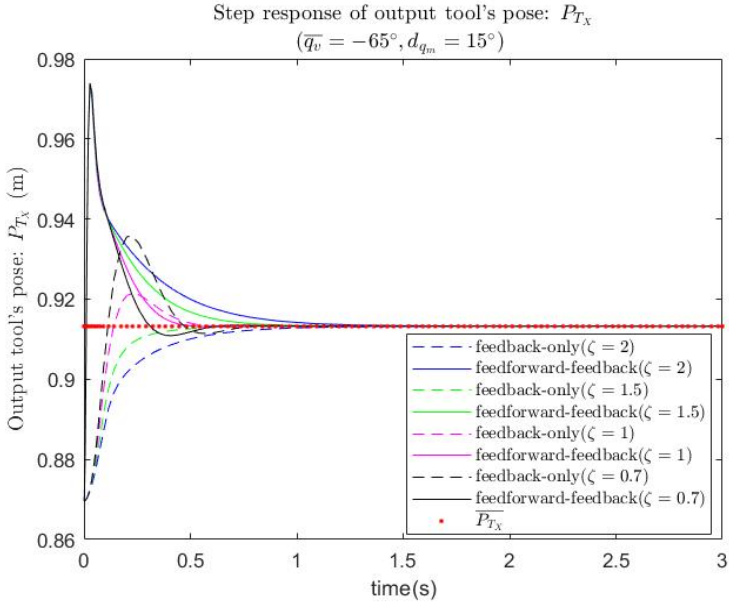

Figure 26. Step responses of $P_{T_X}$. Scenario 2: $\overline{q_v} = -65°$, $d_{q_m} = 15°$.

In summary, the responses from a continuous feedforward-feedback system are more vulnerable to the disturbances especially when the starting position of the tool is far from its target. Although the disturbances will be eliminated as soon as the tool moves inside the camera range of view, the overshoots are more severe if more disturbances are accumulated in the process. In the real-world applications, the feedback-only control solutions are slower than the simulation results as the camera requires extra time, which is not considered in these simulations, to take pictures. Therefore, a feedforward controller, which compensates for the speed limitation of the feedback-only control, becomes indispensable in real manufacturing environments. The issue of the overshoots can be dealt with either by upgrading the camera with a wider range of view or as mentioned previously, the use of a switching algorithm, such as switching from a feedforward to a feedback controller, rather than the use of a continuous feedforward-feedback controller.

**Conclusion**

In this Chapter different sources of uncertainties in the task of positioning control in the automated manufacturing process are introduced. Then, a sequence of control methodologies is proposed. In the first part of this Chapter, we presented movement of a camera in the space to search for an optimal pose, a location in the space where the tool pose can be reached with minimum amount of energy and time duration. In the second part, we discussed a visual servoing architecture, which is applied to eliminate the measurement and dynamic noises occurred in the process of the camera movement. The image averaging technique is used to minimize the image noises by the averaging multiple images. In the last part, we designed the feedback and the feedforward controllers to guide the tool to its target by eliminating the dynamic errors in the tool movement process. Designs of all the control systems have been thoroughly discussed in this Chapter. Our methods for controller design are based on the classical Image Based Visual Servoing (IBVS) technique but are improved on by adding dynamic components to the systems and avoiding the depth estimation as done in the classical methods. Although only one degree of freedom case is discussed in this Chapter, the SISO simulation results have shown great potential of this work for various real-world applications in the automated high-speed manufacturing processes.